%% file: colm2024_conference.tex
\documentclass[dvipsnames]{article} %
\usepackage{colm2024_conference}

\usepackage{booktabs}
\usepackage{enumitem}
\usepackage{wrapfig}
\usepackage{algorithm}
\usepackage{algpseudocode}
\usepackage{graphicx}
\usepackage{subfigure}
\usepackage[misc]{ifsym}

\usepackage{microtype}
\usepackage{amsmath}
\usepackage{colortbl}
\usepackage[utf8]{inputenc}
\usepackage[T1]{fontenc}
\definecolor{lightgray}{rgb}{0.9,0.9,0.9}
\usepackage{caption}
\usepackage{subcaption}
\usepackage{setspace}
\usepackage{url}
\usepackage{multirow}
\usepackage{tabularx}
\usepackage{blindtext}
\usepackage{pgfplots}
\pgfplotsset{compat=1.18} 
\usepackage{tikz}
\usetikzlibrary{er,positioning,bayesnet}
\usepackage{makecell}
\usepackage{tipa}
\usepackage{siunitx}
\usepackage{nicefrac}
\usepackage{listings}
\usepackage[raster,skins, most]{tcolorbox} %
\usepackage{xltabular}
\usepackage{adjustbox}
\usepackage{xurl}
\usepackage{rotating}
\usepackage[normalem]{ulem}
\usepackage[nameinlink]{cleveref}
\usepackage{fontawesome}
\usepackage{pifont}

\definecolor{darkgreen}{RGB}{0,128,0}

\useunder{\uline}{\ul}{}

\input{math_commands.tex}

\newcommand{\ghlink}{https://github.com/Tongyi-MAI/MobileWorld}

\newcommand*\justify{%
  \fontdimen2\font=0.4em%
  \fontdimen3\font=0.2em%
  \fontdimen4\font=0.1em%
  \fontdimen7\font=0.1em%
  \hyphenchar\font=`\-%
}

\renewcommand{\texttt}[1]{%
  \begingroup
  \ttfamily
  \begingroup\lccode`~=`/\lowercase{\endgroup\def~}{/\discretionary{}{}{}}%
  \begingroup\lccode`~=`[\lowercase{\endgroup\def~}{[\discretionary{}{}{}}%
  \begingroup\lccode`~=`.\lowercase{\endgroup\def~}{.\discretionary{}{}{}}%
  \catcode`/=\active\catcode`[=\active\catcode`.=\active
  \justify\scantokens{#1\noexpand}%
  \endgroup
}

\usepackage{makecell}
\usetikzlibrary{tikzmark}
\makeatletter
\newcommand*\myfontsize{%
  \@setfontsize\myfontsize{7}{8}%
}
\makeatother

\definecolor{uclablue}{RGB}{159, 195, 224}

\definecolor{uclagold}{RGB}{255, 240, 180}

\definecolor{aliceblue}{RGB}{255, 238, 241}

\definecolor{cadmiumgreen}{rgb}{0.0, 0.42, 0.24}

\definecolor{myred}{rgb}{0.7, 0.3, 0.0}
\definecolor{myblue}{rgb}{0.2, 0.3, 0.6}
\definecolor{babygreen}{rgb}{0.85, 0.97, 0.85}

\definecolor{purple1}{RGB}{126, 107, 196}
\definecolor{purple2}{RGB}{199, 158, 207}
\definecolor{purple3}{RGB}{214, 200, 255}
\definecolor{purple4}{RGB}{254, 240, 255}

\definecolor{deepblue}{RGB}{48, 58, 82}

\newcommand{\symboletongyi}{\raisebox{0pt}{~\includegraphics[scale=0.012]{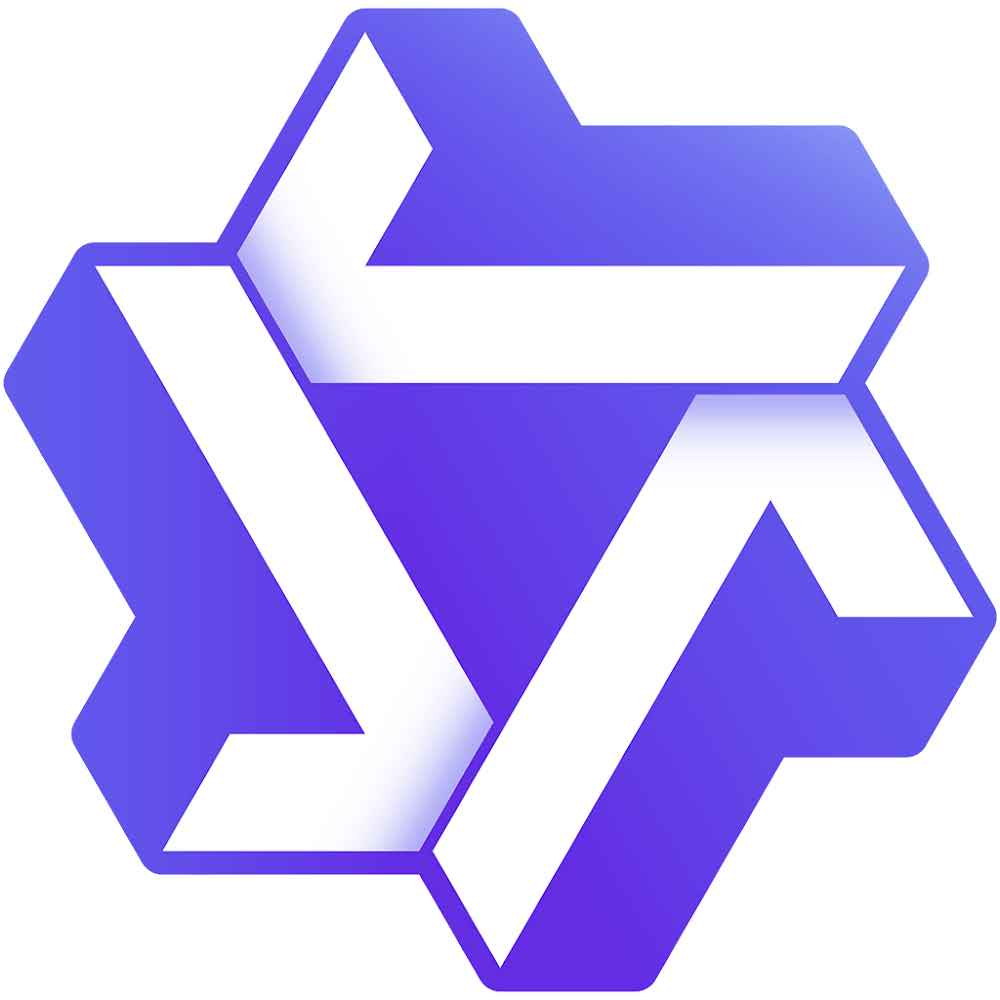}}~}

\definecolor{deepPurple}{HTML}{330066}

\definecolor{uclablue_old}{rgb}{0.15, 0.45, 0.68}
\hypersetup{
    breaklinks,
    citecolor=uclablue_old,
    colorlinks=true,
}

\newtcolorbox{mybox}[2][]
  {colback = black!5!white, colframe = black!75!black, fonttitle = \bfseries,
    colbacktitle = black!100!black, enhanced, before upper={\fontsize{8}{11}\obeyspaces\obeylines\selectfont}, fontupper=\selectfont,
    attach boxed title to top left={yshift=-2.2mm,xshift=4mm},
    title=#2,#1}

\newcommand{\mobileworld}{MobileWorld}

\title{%
    \mobileworld{}: Benchmarking Autonomous Mobile Agents in Agent-User Interactive and MCP-Augmented Environments
}

\author{%
Quyu Kong$^{*,1}$, Xu Zhang$^{*,1}$, Zhenyu Yang$^{2}$, Nolan Gao$^{3}$, Chen Liu$^{1}$, Panrong Tong$^{1}$, Chenglin Cai$^{1}$, Hanzhang Zhou$^{1}$, Jianan Zhang$^{1}$, Liangyu Chen$^{1}$, Zhidan Liu$^{2}$ \\
 Steven HOI$^{1}$, Yue Wang$^{\textrm{1 \Letter}}$
  \\[1em]               %
  {\fontsize{10pt}{11pt}\selectfont          %
$^{1}$Tongyi Lab\symboletongyi, Alibaba Group \quad $^{2}$HKUST (GZ) \quad $^{3}$University of Florida}\\
{\fontsize{9pt}{10pt}\selectfont $^{*}$Equal contribution~~~$\textrm{\Letter}$~\texttt{yue.w@alibaba-inc.com}}
}

\begin{document}

\maketitle

\begin{abstract}

\input{content/abstract.tex}

\end{abstract}

\input{content/intro.tex}

\input{content/related_work}

\input{content/method}

\input{content/experiments}

\input{content/conclusion.tex}

\bibliography{biblio}
\bibliographystyle{colm2024_conference}
\appendix

\input{content/appendix.tex}

\end{document}

%% file: math_commands.tex
\usepackage{amsmath,amsfonts,bm}

\def\eqref#1{equation~\ref{#1}}

\def\1{\bm{1}}

\DeclareMathAlphabet{\mathsfit}{\encodingdefault}{\sfdefault}{m}{sl}
\SetMathAlphabet{\mathsfit}{bold}{\encodingdefault}{\sfdefault}{bx}{n}

%% file: content/abstract.tex
Among existing online mobile-use benchmarks, AndroidWorld has emerged as the dominant benchmark due to its reproducible environment and deterministic evaluation; however, recent agents achieving over 90\% success rates indicate its saturation and motivate the need for a more challenging benchmark.
In addition, its environment lacks key application categories, such as e-commerce and enterprise communication, and does not reflect realistic mobile-use scenarios characterized by vague user instructions and hybrid tool usage.

To bridge these gaps, we introduce \mobileworld{}, a substantially more challenging benchmark designed to reflect real-world usage through 201 tasks across 20 applications.
MobileWorld derives its difficulty from an emphasis on long-horizon, cross-application workflows, requiring nearly twice as many completion steps on average (27.8 vs. 14.3) and featuring a significantly higher proportion of multi-app tasks (62.2\% vs. 9.5\%) than AndroidWorld. 
To overcome the limitations of existing environments, MobileWorld achieves a balance between production-grade utility and reproducible evaluation by utilizing open-source alternatives to industry standards (e.g., Mattermost for Slack). This approach enables a fully observable and controlled environment through source code modification and direct backend database access for precise verification. Furthermore, MobileWorld extends beyond standard GUI manipulation by introducing novel task categories, including agent-user interaction and Model Context Protocol (MCP)-augmented tasks, providing a robust framework for evaluating agents in user-aware, hybrid-tool scenarios.

To facilitate evaluation, we develop a planner-executor agentic framework with extended action spaces to support user interactions and MCP calls. Our results reveal a sharp performance drop compared to AndroidWorld, with the best agentic framework and end-to-end model achieving 51.7\% and 20.9\% success rates, respectively, highlighting ample headroom for future research.
Our analysis further shows that current models struggle significantly with user interaction and MCP calls. By identifying these core research gaps, we offer a strategic roadmap toward next-generation mobile intelligence.

%% file: content/intro.tex
\input{figures/fig-sr}
\section{Introduction}
\label{sec:intro}

GUI agents on smartphones have emerged as a transformative technology for automating mobile tasks, enabling users to delegate complex operations through natural language instructions. With the rapid advancement of Vision-Language Models (VLMs), numerous VLM-based mobile agents have been developed to  navigate mobile interfaces automatically~\citep{rawles2023androidinthewild, deng2024mobile, chen2024spa, yang2025probench}. Reliable benchmarks are therefore essential for measuring their progress. 
Despite the existence of multiple benchmarks, AndroidWorld~\citep{rawles2024androidworld} remains the go-to benchmark for mobile GUI agent evaluation, owing to its reproducible emulator environment and deterministic evaluation.
However, AndroidWorld is approaching saturation --- state-of-the-art agentic frameworks now achieve success rates exceeding 90\% on its leaderboard, fundamentally limiting our ability to distinguish between incremental improvements and genuine breakthroughs in mobile agent capabilities.

Beyond saturation, current mobile agent benchmarks exhibit several fundamental limitations.  \textbf{(1) Existing tasks lack the complexity to reflect real-world mobile usage}. 
Prominent benchmarks like AndroidWorld omits critical application categories—such as e-commerce and enterprise communication—that are central to daily mobile workflows.
Furthermore, prior benchmarks typically feature short-horizon tasks confined within single applications, whereas practical mobile assistance frequently demands long-horizon planning and cross-application workflows. 
\textbf{(2) Existing benchmarks often assume fully-specified instructions}, neglecting realistic scenarios where users provide vague or incomplete requests. In real-world deployments, agents must proactively engage in clarification dialogues to acquire missing information. 
\textbf{(3) Existing  benchmarks overlook the growing importance of external tool invocation}.
The Model Context Protocol (MCP)~\citep{anthropic_mcp_intro} has rapidly emerged as a standardized interface exposing agents to external tools beyond traditional GUI operations. The community is actively debating the integration of GUI operations alongside MCP tools~\citep{wang2025ui}, recognizing that next-generation mobile agents must seamlessly combine interface manipulation with external tool invocations to solve tasks that extend beyond traditional smartphone capabilities. However, no existing benchmark evaluates this hybrid execution paradigm.

To address these limitations, we introduce \mobileworld{}, a substantially more challenging mobile-use online benchmark designed to better reflect real-world mobile usage.

\textbf{Our first contribution} is a substantial increase in task complexity. As illustrated in~\Cref{fig:benchmark_comparison_1}, \mobileworld{} features higher difficulty than AndroidWorld across multiple dimensions: under the agentic framework equipped with GPT-5, \mobileworld{} tasks require on average 27.8 completion steps, nearly twice as many as 14.3 steps required in AndroidWorld.
Moreover, 62.2\% of tasks involve cross-application workflows compared to only 9.5\% in AndroidWorld. These tasks demand challenging capabilities including long-horizon multi-step planning across subtasks, memory retention of past interactions, and precise instruction following. The experimental results clearly reflect this increased difficulty: whereas the best agents achieve success rates exceeding 90\% on AndroidWorld, the top-performing agentic framework reaches only 51.7\% on \mobileworld{}, highlighting substantial room for future improvement.

\input{figures/fig_overview_yue}

\textbf{Our second contribution} is the introduction of \textit{agent-user interaction tasks} that evaluate an agent's ability to handle ambiguous instructions through collaborative dialogue. As shown in the left side of \Cref{fig:system_architecture}, consider the user request: \textit{``Send an email to Kevin with the message ``Hello''.} However, Kevin's email address is not present in the device's Email or Contacts app. In such cases, the agent must recognize the missing information and proactively trigger an \texttt{ask\_user} action to obtain the necessary details, rather than hallucinating or failing silently. Inspired by \citet{yao2024tau}, we employ a \textit{user agent} role-played by a LLM (GPT-4.1 in the example), which holds the key information in its context, and interacts with the GUI agent. This task type tests whether agents can identify the ambiguity from the instruction and engage in clarification dialogues.
22.4\% of all tasks involve agent-user interaction in \mobileworld{}.

\textbf{Our third contribution} is the incorporation of \textit{MCP-augmented tasks} that require hybrid-usage of the GUI navigation and external tool invocations. As illustrated in the right side of \Cref{fig:system_architecture}, an agent is tasked with fetching a GitHub repository's README and sending its summarization via a team collaboration software (Mattermost). 
Instead of navigating the repository through a web browser via a sequence of GUI actions, the agent can invoke an MCP tool to directly retrieve the content. After obtaining the relevant information from the GitHub MCP tool, the agent continues the task through GUI interactions to compose and send the message.
With the rapid adoption of MCP as a standardized protocol for tool integration, evaluating an agent's strategic choice between GUI operations and API-based tools has become essential.
MCP-augmented tasks constitute 19.9\% of all tasks in \mobileworld{}.

\textbf{Our fourth contribution} is an evaluation infrastructure that strikes a balance between reproducible, deterministic verification and a broad coverage of essential real-world applications. As summarized in \Cref{tab:benchmark_comparison}, prior benchmarks for commercial applications --- such as Gmail or YouTube --- suffer from inherent limitations, including mandatory authentication protocols and opaque internal states \citep{chai2025a3, chen2024spa, yang2025probench, yan2025stepguitechnicalreport}. These restrictions frequently necessitate "MLLM-as-a-judge" evaluation, which introduces stochasticity and noise into the assessment process.
We address these challenges by substituting proprietary services with production-grade, self-hosted open-source alternatives (e.g., Mattermost as a Slack alternative). By modifying source code and gaining direct access to backend databases, we establish a fully observable and controlled environment. Our framework employs a multi-faceted validation suite --- comprising textual response verification, backend database inspection, local storage analysis, and application-specific callbacks --- to ensure absolute determinism. Finally, by leveraging Docker-based Android Virtual Devices (AVDs) and snapshot-based state management, we provide a standardized, "push-button" evaluation protocol designed for rigorous community benchmarking.

\textbf{Our fifth contribution} is a systematic empirical study on \mobileworld{} that characterizes the performance frontiers of contemporary mobile agents. To address the unique requirements of our benchmark, we propose a  planner-executor agentic framework as a competitive baseline. We  extend the primitive action space by introducing \texttt{ask\_user} for disambiguation dialogues and \texttt{mcp\_call} for structured tool invocation. This architecture enables the seamless unification of GUI manipulation, multi-turn user interaction, and external tool use within a single closed-loop decision process.
As demonstrated in \Cref{tab:main_res_sr}, our framework  achieves a state-of-the-art (SOTA) success rate of 51.7\% on \mobileworld{}. Crucially, our evaluation of existing end-to-end GUI agents reveals a \textbf{stark capability collapse} on novel task categories: most baseline models score below 10\% on agent-user interaction and near 0\% on MCP-augmented tasks. These findings underscore a fundamental deficiency in current architectures regarding active user interaction and tool-use. Finally, a rigorous error analysis identifies five open research challenges for the community: 
\emph{(i) user ambiguity detection and clarification engagement, (ii) MCP context management, (iii) long-term memory and state checking, (iv) complex logic reasoning} and \emph{(v) spatial-temporal context awareness}.

\input{tables/tab-related-works}

Taken together, MobileWorld aims to serve the community by providing a rigorous and reproducible environment for advancing next-generation mobile agents --- systems that must excel at \textbf{long-horizon reasoning}, \textbf{active user interaction}, and \textbf{MCP tool use}. By characterizing the performance frontiers of contemporary models, we reveal significant capability gaps and identify  core research challenges, offering the community a clear roadmap toward more autonomous and user-aware mobile intelligence.

%% file: figures/fig-sr.tex
\begin{figure*}[h!]
\centering
\includegraphics[width=1\textwidth,page=1]{figures/images/bar_charts-1_1.pdf}
    \caption{
    Compared to AndroidWorld, \mobileworld{} exhibits lower SOTA success rates, longer task horizons, more cross-application tasks, and sharp performance drops for recent models.
    } %
\label{fig:benchmark_comparison_1}
\end{figure*}

%% file: figures/fig_overview_yue.tex
\begin{figure*}[t]
\centering
\includegraphics[width=0.99\textwidth]{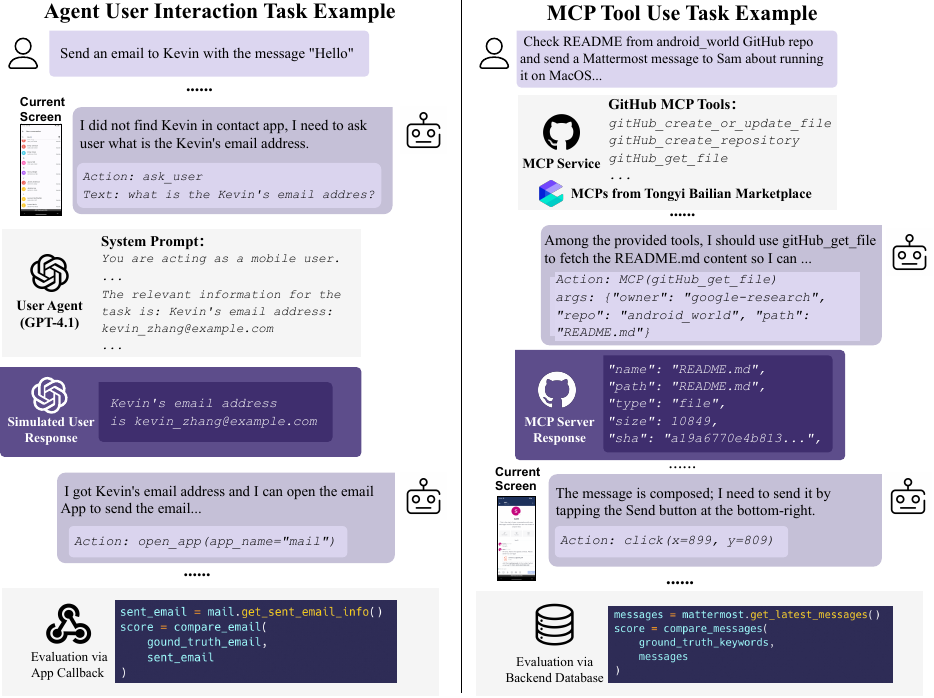}
\caption{
Beyond traditional GUI-only tasks, \mobileworld{} includes agent-user interaction tasks and MCP-augmented tasks, each with distinct deterministic evaluation strategies. \textbf{Left}: An example of an agent-user interaction task, in which the agent must proactively request clarification from a simulated user when encountering incomplete information. A GPT-4.1-based simulated user agent is then triggered to provide the requested information, which is embedded in its system prompt. Task completion is verified through the application's callback cache. \textbf{Right}: An example of an MCP-augmented task, where the agent is initialized with a list of GitHub MCP tools and selects the appropriate tool to retrieve README content from a GitHub repository before completing the task via GUI operations. Task completion is verified through backend database inspection.
}
\label{fig:system_architecture}
\end{figure*}

%% file: tables/tab-related-works.tex
\begin{table}[t]
\centering
\small
\caption{Comparison of online Mobile GUI Agent Benchmarks. 
\mobileworld{} uniquely enables deterministic evaluation even for applications requiring backends (e.g., messaging) by utilizing self-hosted environments. We also introduce novel task paradigms: agent-user interaction and Model Context Protocol (MCP) augmentation.  Backend-Req. Apps: Environment includes third-party apps requiring external backend authentication. (\ding{51} = Supported, \ding{55} = Not Supported).}
\label{tab:benchmark_comparison}
\begin{tabular}{@{}lcccccc@{}}
\toprule
\textbf{Benchmark} & \#Apps & \#Tasks & \makecell{Agent-User \\ Int. Tasks} & \makecell{MCP-Aug. \\ Tasks} & \makecell{Backend-Req.\\Apps} & \makecell{Deterministic \\ Evaluation} \\ 
\midrule

AndroidArena        & 16  & 221  & \textcolor{red}{\ding{55}} & \textcolor{red}{\ding{55}} & \textcolor{darkgreen}{\ding{51}} & \textcolor{red}{\ding{55}} \\
A3                & 20 & 201  & \textcolor{red}{\ding{55}} & \textcolor{red}{\ding{55}} & \textcolor{darkgreen}{\ding{51}} & \textcolor{red}{\ding{55}} \\
Pro-Bench         & 34 & 200  & \textcolor{red}{\ding{55}} & \textcolor{red}{\ding{55}} & \textcolor{darkgreen}{\ding{51}} & \textcolor{red}{\ding{55}} \\
AndroidDaily        & 48 & 235  & \textcolor{red}{\ding{55}} & \textcolor{red}{\ding{55}} & \textcolor{darkgreen}{\ding{51}} & \textcolor{red}{\ding{55}} \\
SPA-Bench         & 66 & 340  & \textcolor{red}{\ding{55}} & \textcolor{red}{\ding{55}} & \textcolor{darkgreen}{\ding{51}} & \textcolor{red}{\ding{55}} \\
\midrule
MobileAgentBench  & 10 & 100  & \textcolor{red}{\ding{55}} & \textcolor{red}{\ding{55}} & \textcolor{red}{\ding{55}}  & \textcolor{darkgreen}{\ding{51}} \\
AndroidLab        & 9  & 138  & \textcolor{red}{\ding{55}} & \textcolor{red}{\ding{55}} & \textcolor{red}{\ding{55}}  & \textcolor{darkgreen}{\ding{51}} \\
AndroidWorld      & 20 & 116  & \textcolor{red}{\ding{55}} & \textcolor{red}{\ding{55}} & \textcolor{red}{\ding{55}}  & \textcolor{darkgreen}{\ding{51}} \\
\midrule
\rowcolor[gray]{0.9} \textbf{\mobileworld{} (Ours)}   & \textbf{20} & \textbf{201}  & \textcolor{darkgreen}{\ding{51}}  & \textcolor{darkgreen}{\ding{51}} & \textcolor{darkgreen}{\ding{51}} & \textcolor{darkgreen}{\ding{51}} \\

\bottomrule
\end{tabular}
\end{table}

%% file: content/related_work.tex
\section{Related Works}

We review existing benchmarks for GUI agents and recent work on agent-user interaction and MCP augmentation, highlighting the gaps that \mobileworld{} addresses.

\paragraph{GUI Agent Benchmarks}
The development of autonomous agents capable of controlling computers has spurred the creation of various benchmarks across different platforms. For desktop environments, benchmarks such as WindowsAgentArena~\citep{bonatti2024windows}, OSWorld~\citep{xie2024osworld}, and WebArena~\citep{zhou2023webarena} evaluate agents on operating system tasks and web browsing scenarios. 
In the mobile domain, several benchmarks have emerged to evaluate Android agents. AndroidWorld~\citep{rawles2024androidworld} introduces a fully functional Android environment with 116 programmatic tasks across 20 real-world apps, featuring dynamic task construction with parameterized natural language instructions. AndroidLab~\citep{xu2025androidlab} provides a systematic framework supporting both LLMs and multimodal models. More recently, Android Agent Arena (A3)~\citep{chai2025a3} addresses some limitations by incorporating 20 widely used third-party commercial apps and 201 tasks with real-time online information retrieval. Other notable efforts include SPA-Bench~\citep{chen2024spa}, ProBench~\citep{yang2025probench}, and AndroidDaily~\citep{yan2025stepguitechnicalreport}, which further expand task coverage and evaluation methodologies.

While these benchmarks have significantly advanced mobile agent research, they face critical limitations summarized in~\Cref{tab:benchmark_comparison}: performance saturation on simpler tasks, the assumption of fully-specified instructions, lack of external tool integration, and a trade-off between realism and deterministic verifiability when using commercial applications.

\paragraph{Agent-User Interaction and MCP-Augmented Benchmarks}
Beyond standard GUI operations, recent work has recognized the importance of evaluating agents on their interaction capabilities and tool usage. $\tau$-bench~\citep{yao2024tau} introduces dynamic conversations between simulated users and agents equipped with domain-specific API tools and policy guidelines, revealing that even state-of-the-art models like GPT-4o succeed on fewer than 50\% of tasks and exhibit low consistency. This highlights the challenge of building agents that can reliably follow rules and handle ambiguous requests. Building on this, $\tau^{2}$-bench~\citep{barres2025tau} models a dual-control environment where both agent and user actively modify shared world states, demonstrating significant performance drops when agents must coordinate with and guide users rather than operate autonomously.

Meanwhile, MCP has emerged as a mechanism for agents to invoke external tools alongside GUI operations. OSWorld-MCP~\citep{jia2025osworld} presents the first benchmark assessing tool invocation, GUI operation, and decision-making in real-world environments, with 158 high-quality tools across seven applications. Their evaluation shows that MCP tools generally improve task success rates (e.g., from 8.3\% to 20.4\% for OpenAI o3), though tool invocation rates remain relatively low at 36.3\%. Similarly, MCPWorld~\citep{yan2025mcpworld} is a benchmark for API, GUI, and hybrid agents using "white-box apps" with source code availability, enabling programmatic verification through dynamic code instrumentation and achieving 75.12\% task completion accuracy.

While these benchmarks advance evaluation of agent-user interaction and tool usage, they operate in domains separate from mobile GUI agents. \mobileworld{} is the first mobile benchmark to integrate both agent-user interaction tasks and MCP-augmented tasks within a unified evaluation framework for practical mobile agent deployment.

%% file: content/method.tex
\section{\mobileworld{}} %
\label{sec:preliminary}
In this section, we formalize the task definition, describe the environment architecture and detail the benchmark construction methodology.

\subsection{Task Definition}

A standard Mobile GUI agent task can be formalized as a \textbf{Partially Observable Markov Decision Process (POMDP)} $(\mathcal{S}, \mathcal{O}, \mathcal{A}, \mathcal{T}, \mathcal{R})$, where $\mathcal{S}$ is the state space representing the mobile environment; $\mathcal{O}$ is the observation space, including a natural language instruction and interface representations such as screenshots; $\mathcal{A}$ is the action space of standard mobile UI operations (see~\Cref{tab:action_space}); $\mathcal{T}: \mathcal{S} \times \mathcal{A} \rightarrow \mathcal{S}$ is the transition function; and $\mathcal{R}: \mathcal{S} \times \mathcal{A} \rightarrow \{0, 1\}$ is a binary reward indicating task completion.

\input{tables/tab-action-space}

\paragraph{GUI-Only Tasks} Following AndroidWorld~\citep{rawles2024androidworld}, GUI-Only tasks comprise task completion and information retrieval, requiring the agent to accomplish user objectives through standard mobile UI operations. Task Completion involves executing a sequence of GUI actions to drive the system into a specified target state (e.g., composing and sending an email), while Information Retrieval requires locating and gathering relevant information from internal mobile data or external sources (e.g., answering questions about today's scheduled events).

\paragraph{Agent-user Interaction Tasks}
Previous mobile GUI agent benchmarks assume that user instructions are clear and unambiguous. However, this assumption rarely holds in real-world scenarios, where users often provide vague or ambiguous instructions, necessitating additional information for GUI agent to narrow down the solution space.
To reflect this practical challenge, we introduce a dedicated set of tasks in which critical information is deliberately omitted from the task goal, rendering successful completion impossible without actively seeking clarification.

\paragraph{MCP-augmented Tasks}
MCP provides a more straightforward way to access a wider range of resources and functionalities beyond standard GUI operations, enabling agents to obtain critical information and complete tasks more effectively~\citep{jia2025osworld}.
However, in real-world scenarios, not all applications are equipped with well-designed MCP interfaces, and the GUI remains the predominant mode of interaction.
Given this co-existence of GUI-based interaction and MCP-enabled tool access, we specifically design a set of tasks augmented with MCP tools.  
These tasks require agents to integrate MCP tool invocations with standard GUI operations to succeed.

\subsection{Environment Development}
\label{sec:env_dev}

As illustrated in~\Cref{fig:system_features}, \mobileworld{} comprises two main components: the host machine running GUI agents and the containerized environment. We detail the key design elements below.

\input{figures/fig-features}

\paragraph{Action Space}
\mobileworld{} supports a hybrid action space that combines standard GUI operations with agent-user interaction and MCP tools invocation. \Cref{tab:action_space} shows the complete action specification.

\textit{GUI Operations.} The core action space includes standard mobile interface manipulations: \texttt{click}, \texttt{double\_tap}, \texttt{long\_press}, \texttt{drag}, \texttt{input\_text}, \texttt{scroll}, and navigation commands (\texttt{navigate\_home}, \texttt{navigate\_back}). Additionally, agents can use \texttt{answer} to respond to information retrieval queries, \texttt{wait} to allow screen updates, and \texttt{status} to signal task completion or infeasibility.

\textit{Agent-User Interaction.} We augment the action space with an \texttt{ask\_user} action that enables agents to proactively request clarification. When invoked, queries are routed to a user agent, implemented with a standard LLM (e.g., GPT-4.1).
It is conditioned on the context information omitted from task instructions. This design enables systematic evaluation of an agent's ability to recognize knowledge boundaries and engage in collaborative dialogue.

\textit{MCP Tool Integration.} Agents can invoke external tools via the \texttt{mcp\_call} action. We curate a collection of popular MCP servers from the Bailian platform\footnote{\url{https://bailian.console.aliyun.com/?tab=mcp\#/mcp-market}}, spanning diverse domains including geospatial navigation (Amap), code repository analysis (GitHub), document processing (Jina), financial data retrieval (StockStar), and scholarly literature fetching (arXiv). These 61 tools enable functionalities that significantly exceed the capabilities of standard GUI operations, such as retrieving commit histories or querying real-time stock data. The complete tool catalog is provided in~\Cref{tab:mcp_tools}.

\input{tables/tab-mcp-tools}

\paragraph{Stable and Reproducible Environment}
\mobileworld{} achieves stability and reproducibility through comprehensive containerization and careful engineering.

\textit{Containerized Architecture.} The entire evaluation environment is encapsulated in Docker-in-Docker containers (right side in \Cref{fig:system_features}), including a rooted Android Virtual Device (AVD), self-hosted application backends, and an API server for orchestration. This design eliminates external dependencies and enables consistent deployment across different host systems.

\textit{Open-Source Applications.} We construct stable, reproducible application environments based on popular open-source projects, including Mattermost (enterprise communication, an open-source alternative to Slack), Mastodon (social media, an open-source alternative to X/Twitter), and Mall4Uni (e-commerce platform). We fork these projects and apply moderate modifications to ensure compatibility within the Docker environment. In addition, self-hosting these applications provides full backend access, enabling precise control over task initialization states and deterministic verification of task outcomes. More details about the open-source applications can be referred in \Cref{sec:app_construction}.

\textit{Snapshot-Based State Management.} To ensure consistent initial conditions across evaluation runs, we employ AVD snapshots that capture the complete device state. Each task execution begins from a predetermined snapshot, guaranteeing reproducible starting conditions.

\paragraph{Task Evaluation}
\label{sec:evaluator}
Deterministic evaluation is essential for reliable benchmarking. We implement multiple complementary verification methods to assess task success:

\textit{Textual Answer Verification.} For information retrieval tasks where the expected output is a specific textual response (e.g., a string or numerical value), we employ pattern matching with regular expressions or exact string comparison against ground-truth answers.

\textit{Backend Database Verification.} For tasks involving self-hosted applications such as Mattermost or Mastodon, we directly query the backend database to verify that expected state changes (e.g., sent messages, created posts) have occurred. This provides ground-truth validation independent of UI state.

\textit{Local Storage Inspection.} Leveraging rooted emulator access, we use Android Device Bridge (ADB) to inspect application-specific local storage and verify internal state. For example, calendar events can be validated by directly examining the Fossify Calendar database, while email drafts can be retrieved from the Mail application's local storage.

\textit{Application Callbacks.} For tasks involving lightweight customized applications, we implement callback APIs that capture intermediate states during execution. These states are persisted and subsequently queried by the evaluator to determine task success.

\Cref{tab:eval_mode} lists some examples of tasks and demonstrate how conduct deterministic evaluation on them.
This verification system ensures that task outcomes are evaluated deterministically, eliminating the evaluation noise inherent in MLLM-as-a-judge approaches used by prior work.

\input{tables/eval_mode_new}

\subsection{Benchmark Construction}

\paragraph{Task Instructions and Scenarios}
To ensure our benchmark reflects diverse and realistic mobile usage scenarios, we curate tasks from a broad spectrum of real-world domains, including e-commerce, social platform, productivity collaboration, on-device system management, and information retrieval. 
All tasks are constructed based on the \mobileworld{} initial system state, which is preloaded by contacts, SMS messages, emails, calendar events, and files.
The preloaded materials are either synthesized by large language models (LLMs) or derived from the internet. 

To increase task complexity, we instruct annotators to design long-horizon tasks that integrate multiple challenging dimensions. These include: (1) combining subgoals into a single objective spanning multiple apps; (2) requiring fine-grained visual recognition (e.g., extracting data from complex PDF layouts); (3) demanding memory retention across steps (e.g., using prior search results in a follow-up email); (4) involving numerical or logical reasoning (e.g., computing total price of cart items); (5) relying on implicit temporal or spatial context (e.g., inferring ``tomorrow'' from the system clock); and (6) enforcing precise instruction following (e.g., responding with only a number). Together, these dimensions help expose capability gaps of GUI agents in realistic mobile scenarios.

\paragraph{Agent-User Interaction Task Construction}
To construct agent-user interaction tasks, we first ask human annotators to write a clear and achievable task goal (e.g., \textit{``Send an email to kevin\_zhang@example.com with the message `Hello.'''}).  
Next, they remove critical information from the instruction --- such as replacing the full email address with only a name like ``Kevin'' --- making the task ambiguous or incomplete.  
This forces GUI agents to actively request clarification to uncover the missing details.
To prevent unintended information leakage, annotators also verify that the device environment does not contain the omitted information (e.g., ensuring ``Kevin'' is not already saved in Contacts app with a linked email).

To respond to GUI agent's clarification requests, we deploy a LLM-simulated \textit{user agent}, a separate LLM configured with the full task goal and relevant background context. 
For those user-interaction tasks, the user agent is injected with the exact information intentionally omitted from the original instruction (e.g., the Kevin's email address in the above example).
In this way, the user agent is able to respond accurately to GUI agent's requests.
In addition, the user agent is instructed not to assist beyond this scope, i.e., if GUI agents ask irrelevant or off-task questions, the user agent should refuse to help. Full prompt for the user agent is provided in the following block.

\input{tables/user_agent_prompt}

\paragraph{MCP-Augmented Task Construction}
Given the selected MCP servers (\Cref{tab:mcp_tools}), annotators are instructed to read each tool's description and sample output. They then create standalone tasks that can be completed using only MCP tools (e.g., ``List all commits in the main branch of the mastodon/mastodon repository from the past week'' via the GitHub MCP).  
These tasks are non-trivial or error-prone via pure-GUI operations. 
Subsequently, annotators extend these into hybrid workflows by appending one or more GUI-based actions that consume the MCP output. For example, after retrieving a commit history via GitHub MCP tool calling, the agent is instructed to compose and send an email summarizing the changes using the native Email app.

\input{tables/benchmark_summary_concise}

\paragraph{Human-in-the-Loop Task Validation}
Once tasks are constructed, annotators manually validate them and ensure that all tasks can be successfully completed by a GUI agent.  
Specifically, annotators initialize the task environment and receive the task goal. They then interact directly with the \mobileworld{} environment, executing actions until they believe the task has been fulfilled.  
Each completed attempt is subsequently evaluated by the evaluator (see~\Cref{sec:evaluator}). A score of 1.0 confirms that the task is solvable under the given conditions. If the score is 0.0, the validator retries the task, up to a maximum of five attempts.  
If the task remains unsolved after five trials, this indicates a potential issue in the task initialization, instruction clarity, or evaluation criteria. Such tasks are flagged and returned to the design phase for revision and reimplementation.

\subsection{Data Statistics}

We summarize key statistics of \mobileworld{} in~\Cref{fig:scenario_distribution} and~\Cref{tab:statistics}, highlighting four core dimensions. (1) Tasks span diverse domains, including communication, productivity, and navigation. Approximately 95\% of tasks involve third-party applications, aligning with authentic
mobile usage patterns and ensuring the benchmark's real-world relevance. 
(2) The benchmark emphasizes cross-app complexity: 62.2\% of tasks require coordination between multiple apps, and 12.4\% involve three or more. 
(3) Agent-user interaction tasks and MCP-augmented tasks account for 42.3\% of the dataset. These tasks go beyond pure GUI navigation by requiring dynamic user engagement and external tool use.  
(4) Multiple verification methods are used to assess task success: 47.3\% of tasks rely on self-hosted database verification, and 36.8\% use local storage inspection, highlighting the importance of apps with fully controlled backends.

%% file: tables/tab-action-space.tex
\definecolor{categoryrow}{gray}{0.90}

\begin{table}[t]
\centering
\small
\caption{Complete action space supported by \mobileworld{}.}
\label{tab:action_space}
\renewcommand{\arraystretch}{1.25}
\begin{tabularx}{\textwidth}{@{}>{\raggedright\arraybackslash}p{2.5cm} >{\raggedright\arraybackslash}p{4.5cm} >{\raggedright\arraybackslash}X@{}}
\toprule
\textbf{Action} & \textbf{Parameters} & \textbf{Description} \\
\midrule
\rowcolor{categoryrow}
\multicolumn{3}{@{}l}{\textit{GUI Operations}} \\
\texttt{click} & \texttt{x}, \texttt{y} & Tap at the specified coordinates \\
\texttt{double\_tap} & \texttt{x}, \texttt{y} & Double-tap at the specified coordinates \\
\texttt{long\_press} & \texttt{x}, \texttt{y} & Long-press at the specified coordinates \\
\texttt{drag} & \texttt{start\_x}, \texttt{start\_y}, \texttt{end\_x}, \texttt{end\_y} & Drag from start to end coordinates \\
\texttt{input\_text} & \texttt{text} & Type text into the focused field \\
\texttt{scroll} & \texttt{direction} & Scroll in the specified direction (up/down/left/right) \\
\midrule
\rowcolor{categoryrow}
\multicolumn{3}{@{}l}{\textit{Navigation}} \\
\texttt{navigate\_home} & --- & Return to the home screen \\
\texttt{navigate\_back} & --- & Navigate to the previous screen \\
\texttt{keyboard\_enter} & --- & Press the enter key \\
\texttt{wait} & --- & Wait for screen content to update \\
\midrule
\rowcolor{categoryrow}
\multicolumn{3}{@{}l}{\textit{Task Control}} \\
\texttt{answer} & \texttt{text} & Provide a textual response to the user (for IR tasks) \\
\texttt{status} & \texttt{goal\_status} & Mark task as \texttt{complete} or \texttt{infeasible} \\
\midrule
\rowcolor{categoryrow}
\multicolumn{3}{@{}l}{\textit{Extended Actions}} \\
\texttt{ask\_user} & \texttt{text} & Request clarification from the user \\
\texttt{mcp\_call} & \texttt{tool\_name}, \texttt{params} & Invoke an MCP tool with specified parameters \\
\bottomrule
\end{tabularx}
\end{table}

%% file: figures/fig-features.tex
\begin{figure*}[t]
\centering
\includegraphics[width=0.9\textwidth]{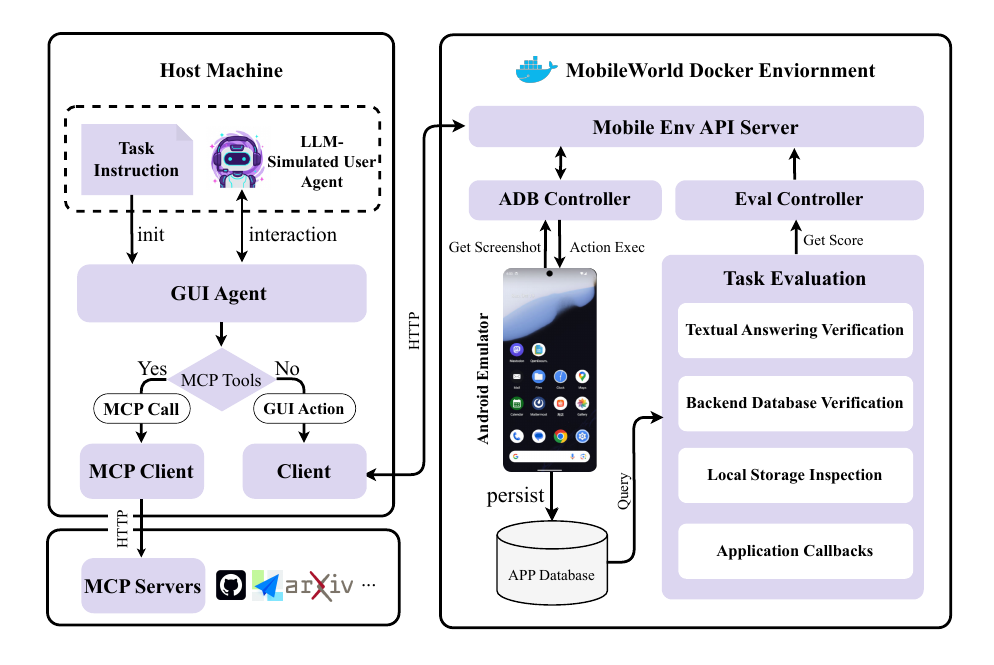}
\caption{The system architecture of \mobileworld{} consists of two main components. \textbf{Left}: the host machine is where GUI agents receive task instructions and optionally interact with users for clarification, then choose between GUI actions or MCP tool calls to complete tasks. \textbf{Right}: the docker environment contains an isolated Android ecosystem with emulators, self-hosted app backends, and an evaluator that verifies task completion through text matching, backend database, local storage, and app callbacks.} %
\label{fig:system_features}
\end{figure*}

%% file: tables/tab-mcp-tools.tex
\begin{table}[t]
\centering
\small
\caption{MCP tools integrated in \mobileworld{}. These tools provide advanced capabilities beyond standard GUI operations, enabling agents to complete complex cross-domain tasks.}
\label{tab:mcp_tools}
\begin{tabular}{lcp{10cm}}
\toprule
\textbf{MCP Server} & \textbf{\# Tools} & \textbf{Description} \\
\midrule
Amap Maps & 15 & Provides comprehensive geospatial services including geocoding, reverse geocoding, IP location, weather queries, and multi-modal route planning (cycling, walking, driving, public transit). Supports distance measurement and location-based search (keyword, nearby, and detail queries). \\
\midrule
GitHub & 26 & Official GitHub integration offering advanced automation and interaction capabilities for developers. Enables repository querying, commit history retrieval, issue tracking, and code analysis. \\
\midrule
Jina AI & 3 & Document processing and search capabilities powered by Jina AI, supporting multi-modal content analysis and retrieval tasks. \\
\midrule
Stockstar & 16 & Financial intelligence service providing comprehensive data for A-share and Hong Kong stocks. Includes fundamental data, derived metrics, financial analysis, and business model evaluation. \\
\midrule
arXiv & 4 & Academic paper search and retrieval from the arXiv repository, enabling agents to access and process scholarly literature. \\
\bottomrule
\end{tabular}
\end{table}

%% file: tables/eval_mode_new.tex
\begin{table*}[t]
\centering
\small
\caption{The example tasks with their evaluation modes and corresponding verification logics.}
\label{tab:eval_mode}
\begin{tabular}{
    >{\raggedright\arraybackslash}p{6cm}
    >{\raggedright\arraybackslash}p{3cm}
    >{\raggedright\arraybackslash}p{5.5cm}
}
\toprule
 \textbf{Task} & \textbf{Eval Mode} & \textbf{Eval Logic} \\
\midrule
 I want to drive to Tianjin. Please check the driving distance in kilometers. Response only one integer number. No other text. & Textual Answering Verification & Compare the agent answer text with the ground-truth distance. \\
\midrule
 Reply to the toot of gourmet user about Greek food Moussaka, and the reply content should be `Nice sharing, i love it'. & Backend Database Verification & Fetch the reply content from the Mastodon backend database and compare with the ground-truth content. \\
\midrule
Set a weekend alarm for 8:25 a.m. with the ringtone ``beebeep'' and vibration off. & Local Storage Inspection & Leverage adb command to query the local storage of Alarm app. \\
\midrule
 I want to remove some electronic products in the shopping cart of the TaoDian app. & Application Callbacks & Implement callback APIs that capture the cart item changes in Taodiao app and persist in a local file for evaluation. \\
\bottomrule
\end{tabular}
\end{table*}

%% file: tables/user_agent_prompt.tex
\begin{tcolorbox}[
title=User Agent System Prompt]
\label{box:user_sys_prompt_example}
\par\textbf{\# System Prompt:}
You are acting as a mobile phone user.
An mobile GUI agent is executing a task on your phone.
The task goal is: \{self.goal\}.
You need to answer questions from the mobile GUI agent.
The relevant information for the task is: \{self.relevant\_information\}.
If the question is not related to the task or no more task-related information is available, you need to refuse to answer in a polite manner.
DO NOT make up any information. You can ONLY give the answer based on the relevant information and the task goal.
Today is 2025-10-16, Thursday. If the question is about the date, you need to answer the correct date based on the current date.
\end{tcolorbox}

%% file: tables/benchmark_summary_concise.tex
\begin{table}[t]
\centering
\small
\begin{minipage}{0.55\textwidth}
    \centering
    \includegraphics[width=\linewidth]{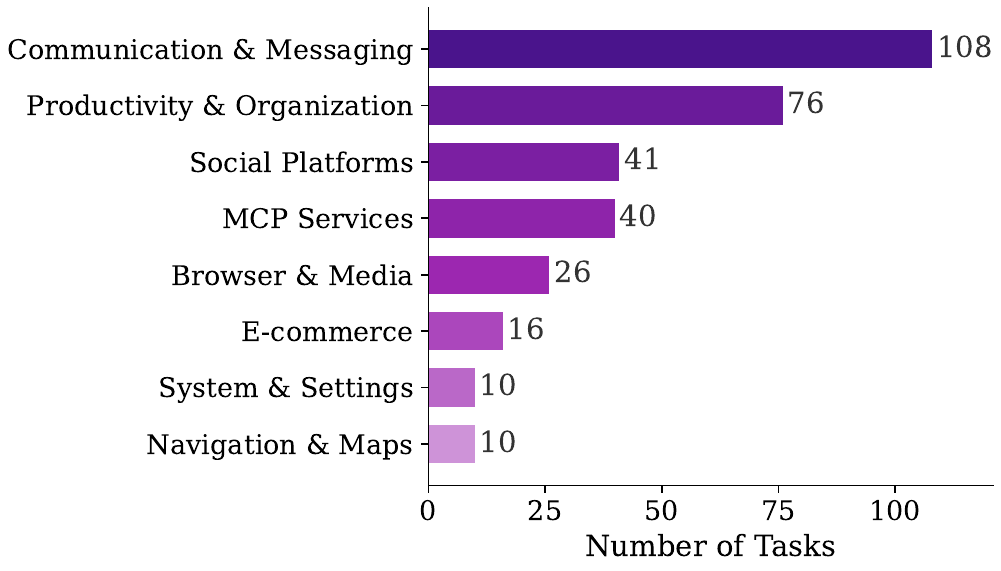}
    \captionof{figure}{
    Scenario Distribution. The benchmark predominantly features third-party applications (95\%), with system apps comprising the remaining 5\% of tasks.
    }
    \label{fig:scenario_distribution}
\end{minipage}
\hfill
\begin{minipage}{0.4\textwidth}
    \centering
    \captionof{table}{Key statistics in \mobileworld{}.}
    \vspace{-1em}
    \label{tab:statistics}
    \begin{tabular}{@{}lc@{}}
    \toprule
    \textbf{Statistic} & \textbf{Number} \\
    \midrule
    Total tasks & 201 (100\%) \\
    \midrule
    \textit{Category Breakdown} & \\
    -- GUI-Only Tasks  & 116 (57.7\%)\\
    -- Agent-User Int. Tasks & 45 (22.4\%)\\ 
    -- MCP-Aug. Tasks & 40 (19.9\%)\\
    \midrule
    \textit{App Complexity} & \\
    -- Single App Tasks & 76 (37.8\%)\\
    -- Two Apps Tasks & 100 (49.8\%) \\
    -- $\geq$ Three Apps Tasks & 25 (12.4\%) \\
    \midrule
    \textit{Evaluation Mode} &  \\
    -- App Callbacks & 10 (5.0\%) \\
    -- Textual Matching & 22 (10.9\%) \\
    -- Storage Inspection & 74 (36.8\%) \\
    -- DB Verification & 95 (47.3\%) \\
    \midrule
    Total MCP Servers & 5 \\
    Total MCP Tools & 64 \\ 
    \bottomrule
    \end{tabular}
\end{minipage}

\end{table}

%% file: content/experiments.tex
\section{Experiments}
In this section, we conduct comprehensive experiments to evaluate state-of-the-art models on \mobileworld{} and analyze their capabilities across different task categories.

\subsection{Agentic Framework for \mobileworld{}}
\label{sec:agent_framework}
To accommodate with the new features in \mobileworld{}, we develop a planner-executor agentic framework. 
It consists of two core components: a \textit{planner} and a \textit{grounding executor}~\citep{jt_guiagent2025}. The planner is responsible for interpreting the task instruction, analyzing the current screenshot along with the interaction history, and deciding the next action, such as \texttt{click}, \texttt{swipe}, \texttt{wait} or \texttt{type}. When the planned action involves clicking (e.g., \texttt{click} or \texttt{long\_press}), the planner does not output coordinates directly; instead, it generates a natural language description of the target UI element (e.g., ``the `Send' button at the bottom right corner'' or ``the email thread titled `Lunch Meeting'.''). This description is then passed to the grounding executor, which takes the current screenshot as input and predicts the precise pixel coordinates to perform the click or long press.
In this way, any general-purpose VLMs can be integrated into our framework, without requiring them to possess pixel-level grounding capabilities.

To support the unique requirements of \mobileworld{}, we extend the action space with two special operations: \texttt{ask\_user} and \texttt{mcp\_call}. When the planner issues an \texttt{ask\_user} action, the system forwards the query to the llm-simulated user agent; the user agent's textual response --- along with the current screenshot --- is appended to the interaction history. 
For MCP-augmented tasks, we preload the specifications of the involved MCP tools into the system prompt.
When the planner triggers \texttt{mcp\_call}, it specifies the MCP tool name and fills in the required parameters.
The MCP invocation request is then sent to the MCP server, which returns a structured output. This result, along with the current screenshot, is injected into the interaction history as the system's observation for that step. This design enables seamless integration of external tool invocation, user clarification, and GUI operation within a unified decision loop.
The details of our agentic framework can be referred in \Cref{sec:agent_framework_appendix}.

\subsection{Metrics}\label{ssec:metrics}
We define the following metrics to measure the effectiveness and efficiency of GUI agents on \mobileworld{}.
\textbf{Success Rate (SR)} measures the proportion of tasks successfully completed by the agent. For each task \(i\) in the \mobileworld{}, a binary score \(s_i \in \{0, 1\}\) is assigned, where \(s_i = 1\) if the task objective is fully achieved and \(s_i = 0\) otherwise. The success rate is computed as:
\begin{equation}
    \text{SR} = \frac{1}{N} \sum_{i=1}^{N} s_i.
\end{equation}
We also report success rates for specific task categories. The \textbf{GUI-Only SR} is computed over GUI-Only tasks, the \textbf{Interaction SR} is computed over agent-user interaction tasks in \(\mathcal{I}_{\text{interact}}\), and the \textbf{MCP SR} is computed over MCP-augmented tasks in \(\mathcal{I}_{\text{MCP}}\):
\begin{equation}
    \text{SR}_{\text{GUI-Only}} = \frac{1}{|\mathcal{I}_{\text{GUI-Only}}|} \sum_{i \in \mathcal{I}_{\text{GUI-Only}}} s_i, \quad
    \text{SR}_{\text{interact}} = \frac{1}{|\mathcal{I}_{\text{interact}}|} \sum_{i \in \mathcal{I}_{\text{interact}}} s_i, \quad
    \text{SR}_{\text{MCP}} = \frac{1}{|\mathcal{I}_{\text{MCP}}|} \sum_{i \in \mathcal{I}_{\text{MCP}}} s_i.
\end{equation}

\textbf{Average Completion Steps (Ave. Steps)} denotes the average number of action steps taken across all execution trajectories. Let \(t_i\) be the number of steps in the trajectory for task \(i\) (including both successful and failed episodes). The average completion steps is defined as:
\begin{equation}
    \text{Ave. Steps} = \frac{1}{N} \sum_{i=1}^{N} t_i.
\end{equation}
Under comparable success rates, a lower value reflects greater execution efficiency.

\textbf{Average User Queries (Ave. Queries)} measures the average number of \texttt{ask\_user} actions invoked by the agent across agent-user interaction tasks. Let \(c_i\) denote the number of \texttt{ask\_user} actions for task \(i \in \mathcal{I}_{\text{interact}}\). This metric is defined as:
\begin{equation}
    \text{Ave. Queries} = \frac{1}{|\mathcal{I}_{\text{interact}}|} \sum_{i \in \mathcal{I}_{\text{interact}}} c_i.
\end{equation}
This metric reflects how frequently agents seek user clarification, with optimal agents achieving task success while minimizing unnecessary queries.

\textbf{User Interaction Quality (UIQ)} evaluates both the effectiveness and efficiency of the agent's \texttt{ask\_user} invocations. Let \(\mathcal{I}_{\text{interact}}\) denote the set of agent-user interaction task indices, and let \(\mathcal{I}_{\text{triggered}}\) denote the set of non-interaction task indices where the agent invoked at least one \texttt{ask\_user} action. For each task \(i\), let \(c_i\) denote the number of \texttt{ask\_user} actions invoked and \(s_i \in \{0, 1\}\) the task success indicator. The quality score for task \(i \in \mathcal{I}_{\text{interact}}\) is computed as:
\begin{equation}
    q_i = \begin{cases}
        \frac{s_i}{c_i} & \text{if } c_i > 0, \\
        0 & \text{if } c_i = 0.
    \end{cases}
\end{equation}
The overall UIQ score is then defined as:
\begin{equation}
    \text{UIQ} = \frac{\sum_{i \in \mathcal{I}_{\text{interact}}} q_i}{|\mathcal{I}_{\text{interact}}| + |\mathcal{I}_{\text{triggered}}|}.
\end{equation}
This metric rewards agents that successfully complete interaction tasks with fewer queries, while penalizing agents that fail to seek necessary clarification or unnecessarily invoke \texttt{ask\_user} on non-interaction tasks. Higher UIQ indicates the agent appropriately recognizes when user clarification is needed and efficiently resolves ambiguities.

\textbf{Average MCP Tool Calls (Ave. MCP Calls)} measures the average number of MCP tool invocations by the agent across MCP-augmented tasks. Let \(m_i\) denote the number of MCP tool calls for task \(i \in \mathcal{I}_{\text{MCP}}\). This metric is defined as:
\begin{equation}
    \text{Ave. MCP Calls} = \frac{1}{|\mathcal{I}_{\text{MCP}}|} \sum_{i \in \mathcal{I}_{\text{MCP}}} m_i.
\end{equation}
This metric indicates the agent's propensity to leverage external tools, with effective agents appropriately integrating MCP tool access into their task execution workflows.

\subsection{Experimental Setup}

We evaluate a suite of state-of-the-art Large Multimodal Models (LMMs), including Qwen3-VL-235B-A22B-Instruct~\citep{qwen3vl2025}, Qwen3-VL-32B-Instruct~\citep{qwen3vl2025}, GUI-Owl-32B~\citep{ye2025mobile}, UI-Venus-72B~\citep{ui-venus},  Doubao-1.5-UI-TARS~\citep{ui-tars-15-seed},  GELab-Zero~\citep{gelab_zero_2025}, OpenAI GPT-5~\citep{gpt5}, Claude-4.5-Sonnet~\citep{anthropicClaudeSonnet}, and Gemini-3-Pro~\citep{comanici2025gemini}. 
To enable consistent comparison of performance across models, we integrate general-purpose LMMs, such as GPT-5 and Claude-4.5-Sonnet, into the planner-executor agentic framework introduced in~\Cref{sec:agent_framework}. In this setup, LMMs serve as the planner, while UI-Ins-7B~\citep{chen2025ui}, a state-of-the-art grounding model, acts as the executor to carry out coordinate-based actions.
For end-to-end GUI agent models, inlcuding Doubao-1.5-UI-TARS, GUI-Owl, UI-Venus, GELab-Zero and Qwen3-VL, we employ their official implementations and adapt them to the \mobileworld{} environment.

All models are evaluated with a maximum of 50 steps using metrics defined in~\Cref{ssec:metrics}.
Agentic frameworks and Qwen3-VL models are evaluated on all task categories. MCP tools are integrated into the system prompt following the tool call format for agentic frameworks, while Qwen3-VL models adopt the OSWorld-MCP format~\citep{jia2025osworld}. Additionally, we augment the Qwen3-VL action space with \texttt{ask\_user} action to enable evaluation on agent-user interaction tasks.
Other end-to-end models are evaluated only on compatible tasks: MCP-augmented tasks are excluded due to lack of tool invocation support, and agent-user interaction tasks are excluded for models without user query actions (e.g., GUI-Owl, UI-Venus).

Open-source models are deployed using vLLM~\citep{kwon2023efficient} on a server with 8$\times$NVIDIA H20 GPUs. Proprietary models including GPT-5, Claude-4.5-Sonnet, Gemini-3-Pro, and Doubao-1.5-UI-TARS are accessed through their official API. The temperature is set to 0.0. The user agent for agent-user interaction tasks is implemented using GPT-4.1.

\input{tables/main_res_1}

\subsection{Main Results}
\label{sec:main_results}

\Cref{tab:main_res_sr} presents the success rate comparison across all evaluated models under different task categories. Several key observations emerge from our results.

First, \mobileworld{} presents a significant challenge even for state-of-the-art models. The best-performing model, GPT-5 + UI-Ins-7B, achieves only 51.7\% overall success rate, substantially lower than the 90\%+ success rates reported on AndroidWorld~\citep{rawles2024androidworld}. This confirms that our benchmark effectively addresses the saturation problem in existing mobile agent evaluation.

Second, there exists a substantial performance gap between agentic frameworks and end-to-end GUI-specialized models. Agentic frameworks leveraging frontier LLMs (e.g., GPT-5, Claude-4.5-Sonnet and Gemini-3-Pro) consistently outperform end-to-end models by a large margin.
The best agentic model achieves an overall success rate of 51.7\%, compared to only 20.9\% for the best end-to-end model (Doubao-1.5-UI-TARS).  
This gap stems from two key limitations of current end-to-end approaches: (1) insufficient capability in handling the complex reasoning and cross-app coordination required by \mobileworld{}, and (2) a lack of support for agent-user interaction and MCP tool invocation.

Third, agent-user interaction and MCP-augmented tasks prove particularly challenging and reveal critical capability gaps. On agent-user interaction tasks, Qwen3-VL and GELab-Zero achieve scores below 10\%, while Doubao-1.5-UI-TARS achieves only 32.4\%.
This highlights the difficulties in recognizing when user clarification is needed. In contrast, GPT-5 demonstrates strong performance (62.2\%) on these tasks, suggesting that advanced reasoning capabilities are essential for effective human-agent collaboration. Similarly, MCP-augmented tasks expose fundamental limitations in tool integration. Qwen3-VL is unable to properly utilize MCP tools when required to coordinate their invocation with GUI actions. On the other hand, among agentic frameworks, the best performance reaches only 51.6\% (GPT-5). This reveals that current models struggle to effectively orchestrate between GUI actions and external tool invocations.

\input{tables/main_res_2}
\subsection{In-depth Analysis}
\Cref{tab:main_res_detailed} presents a comprehensive analysis of agent behavior beyond task success rates. Several notable patterns emerge from examining execution efficiency and interaction quality.

\paragraph{Execution Efficiency} Among all evaluated models, Gemini-3-Pro + UI-Ins-7B achieves the lowest average completion steps (24.2), followed by Claude-4.5-Sonnet + UI-Ins-7B (26.6) and GPT-5 + UI-Ins-7B (27.8). Interestingly, Doubao-1.5-UI-TARS and GUI-Owl-7B also exhibit relatively low step counts (20.9 and 20.6 respectively), but this is largely attributable to their lower success rates---these models often terminate early due to failure rather than efficient task completion. In contrast, UI-Venus-72B requires the highest average steps (34.2) while achieving only 10.4\% success rate, indicating inefficient behavior.

\paragraph{User Interaction Quality} The Average Queries and UIQ metrics jointly reveal how models manage agent-user interaction tasks. GPT-5 demonstrates the most effective clarification behavior, with an average of 1.11 queries per interaction task --- corresponding to its highest user-interaction success rate (62.2\%) and UIQ (0.40). Claude-4.5-Sonnet follows, averaging 0.76 queries and achieving a UIQ of 0.25. In contrast, Gemini-3-Pro got a lower UIQ (0.19) and asks very few queries on average (0.36), even though it performs well on GUI-only tasks (55.6\%). This suggests it under-utilizes user clarification even when beneficial.
Notably, UIQ scores are consistently lower than User-Interaction SR across all models, indicating the presence of unnecessary \texttt{ask\_user} operations that inflate query counts without contributing to task success. In particular for Doubao-1.5-UI-TARS, a User-Interaction SR of 32.4\% was achieved with 1.22 average queries, but the model obtains a lower UIQ of only 0.13. This gap suggests that Doubao-1.5-UI-TARS frequently issues redundant or ineffective clarification requests.

\paragraph{MCP Tool Integration} For MCP-augmented tasks, Gemini-3-Pro leads in average tool invocations (2.63 calls per task), followed by GPT-5 (2.23) and Claude-4.5-Sonnet (1.91). This higher tool utilization correlates with stronger MCP SR performance, indicating that successful MCP task completion requires agents to actively and appropriately leverage external tools. 
The average MCP Tool calls of Qwen3-VL  (2.32 $\sim$ 3.84) is comparable to that of the aforementioned models. 
However, due to inaccuracies in generated tool names and arguments, a large proportion of tool invocation failed, leading to a significantly lower MCP success rate~(0.0 $\sim$ 5.4).
Other end-to-end models universally fail to invoke MCP tools, as they lack the architectural support for tool integration.

These findings highlight that effective mobile agents require not only strong task completion capabilities but also appropriate utilization of interactive clarification and external tool integration, where current end-to-end approaches show significant room for improvement.

\paragraph{Completion Step Comparsion with AndroidWorld}
We compared the task completion steps with 50 max steps under the GPT-5 + UI-Ins-7B agentic framework on \mobileworld{} and AndroidWorld, respectively.
\input{figures/fig-step-distribution}
As shown in~\Cref{fig:benchmark_comparison}, \mobileworld{} exhibits a significantly higher complexity compared to AndroidWorld in terms of task length and step distribution. While AndroidWorld tasks are predominantly completed within 15 steps (with an average of 14.3 steps), \mobileworld{} tasks require substantially more actions, with 27.8 average steps. The distribution of \mobileworld{} tasks is notably right-skewed, with a large proportion requiring over 20 steps, reflecting its focus on long-horizon workflows involving GUI interactions, user clarification, and MCP tool calls.

\subsection{Research Challenge Analysis}

\input{figures/fig-case_study}

To better understand the limitations of current GUI agents, we manually inspect failed task trajectories across evaluated models and categorize representative cases. This analysis reveals several challenges that current agents must overcome to achieve reliable real-world performance.

\paragraph{Challenge 1: Ambiguity Detection and User Engagement}
When the model operates without the ability to query the user for clarification, it often generates plausible-sounding but factually incorrect or unsupported responses. This hallucination arises because the model attempts to infer missing or ambiguous information on its own, rather than seeking confirmation. 
In the example shown in \Cref{fig:hallucination_example}, the task instruction is: \textit{``I want to drive to Tianjin, China from my hometown. Please check the driving distance.''} The GUI agent opens Google Maps and correctly inputs the destination (Tianjin), but it hallucinates ``Shanghai'' as the hometown without asking the user to specify the departure city. This incorrect assumption leads to a wrong distance calculation and ultimately causes task failure.
Allowing interactive clarification significantly reduces these risks by grounding the model's actions in verified user intent.

\paragraph{Challenge 2: MCP Tool Descriptions and Output Management}
External tools invoked via MCP may return excessively long outputs that overwhelm the agent's context window. As illustrated in \Cref{fig:mcp_overflow_example}, when extracting specific benchmark scores from an academic PDF to email a comparison summary, the MCP tool returns the entire 20k-token document as raw text. This floods the context with irrelevant content, preventing the agent from locating target information and causing incorrect extractions or downstream failures. Effective MCP integration requires content-aware retrieval strategies and context management mechanisms~\citep{jones2024codeexecution}.

\paragraph{Challenge 3: Long-Term Memory and State Tracking}
The GUI agent struggles to maintain awareness of actions it has already performed without long-term memory tracking mechanism. As illustrated in \Cref{fig:rename_loop_example}, the agent is instructed to rename all files in the Downloads folder with the prefix \texttt{bid\_}, ordering them by creation date and renaming them uniformly as \texttt{bid\_{index}.{extension}}. However, after successfully renaming several files, the agent forgot the processed files. In subsequent steps, it attempts to rename these renamed files again. This leads to repeated, conflicting modifications and ultimately results in an incorrect final state. The absence of a reliable memory mechanism to track completed subtasks causes the agent to fall into a destructive loop, highlighting a critical gap in its ability to handle multi-step operations in real-world environments.

\paragraph{Challenge 4: Complex Logic Reasoning}
The GUI agent exhibits limited capability in tasks requiring multi-step logical reasoning and accurate numerical calculation. For instance, in the task ``Find the three most expensive items in the TaoDian app shopping cart and calculate their total price,'' the agent must first scan all items in the cart, maintain a running record of the top three highest-priced products, and then sum their prices precisely. While the model attempts this process and outputs a final number, the result is incorrect, either because it fails to correctly identify the most expensive items (e.g., due to misreading prices or overlooking items) or because it makes arithmetic errors during summation. This reflects a broader weakness in handling structured reasoning and exact calculation, which are essential for reliable performance in real-world e-commerce or financial tasks.

\paragraph{Challenge 5: Temporal-Spatial Context Awareness}
The model generally lacks awareness of real-world time and location unless these are explicitly stated in the prompt. In the task ``I've received a lunch invitation via text message; please reply `OK' and schedule a lunch event tomorrow'' the agent is expected to determine today's date --- by observing system UI elements such as the desktop clock --- and create a calendar entry for an appropriate upcoming lunchtime. Most tested agents either hallucinate an arbitrary date or fail to consult time. This inability to ground actions in the actual temporal context leads to incorrectly scheduled events and highlights a critical gap in the agent's situational awareness. Reliable access to the real-time information on the device is essential for performing scheduling tasks accurately.

%% file: tables/main_res_1.tex
\begin{table}[t]
\caption{Success rate (\%) comparison of state-of-the-art mobile GUI agent models on \mobileworld{} under maximum 50 steps. We report overall SR and breakdown by task category:  GUI-Only tasks, agent-user interaction tasks, and MCP-augmented tasks. The number of tasks is indicated in the corresponding column. \textbf{Bold} indicates the best result and \underline{underline} indicates the second best.}
\label{tab:main_res_sr}
\centering
\small
\begin{tabular}{@{}m{2.2cm}l|c|ccc@{}}
\toprule
\textbf{Category} & \textbf{Model} & \textbf{Overall} & \textbf{GUI-Only (116)} & \textbf{User-Int. (45)} & \textbf{MCP (40)} \\
\midrule
\multirow{3}{=}{\centering\textit{Agentic\\Framework}} 
& Claude-4.5-Sonnet + UI-Ins-7B   & \underline{43.8} & 47.8 & \underline{37.8} & \underline{50.0} \\
& Gemini-3-Pro + UI-Ins-7B        & 46.3 & \textbf{55.6} & 24.4 & 48.6 \\
& GPT-5 + UI-Ins-7B          & \textbf{51.7} & \underline{54.0} & \textbf{62.2} & \textbf{51.6} \\
\midrule
\multirow{9}{=}{\centering\textit{End-to-End \\ Model}} 
& GUI-Owl-7B                   & 4.5 & 7.7 & - & - \\
& GUI-Owl-32B                  & 5.5 & 8.5 & - & - \\
& UI-Venus-7B                  & 5.5 & 8.5 & - & - \\
& UI-Venus-72B                 & 10.4 & \underline{16.4} & - & - \\
& Qwen3-VL-8B                  & 5.5 & 9.4 & 0.0 & 0.0 \\
& Qwen3-VL-32B                 & 9.0 & 11.9 & \underline{6.7} & \underline{2.7} \\
& Qwen3-VL-235B-A22B           & 9.5 & 12.8 & 4.4 & \textbf{5.4} \\
& GELab-Zero-4B           & \underline{10.9} & 16.1 & \underline{6.7} & - \\
& Doubao-1.5-UI-TARS     & \textbf{20.9} & \textbf{26.3} & \textbf{32.4} & - \\
\bottomrule
\end{tabular}
\end{table}

%% file: tables/main_res_2.tex
\begin{table}[t]
\caption{Detailed metrics comparison on \mobileworld{} under maximum 50 steps. We report Success Rate (SR), Average Completion Steps (Steps), Average User Queries (Queries), User Interaction Quality (UIQ), and Average MCP Calls (MCP). \textbf{Bold} indicates the best result and \underline{underline} indicates the second best.}
\label{tab:main_res_detailed}
\centering
\small
\begin{tabular}{@{}m{2.2cm}l|c|cccc@{}}
\toprule
\textbf{Category} & \textbf{Model} & \textbf{SR (\%)} & \textbf{Steps} & \textbf{Queries} & \textbf{UIQ} & \textbf{MCP} \\
\midrule
\multirow{3}{=}{\centering\textit{Agentic\\Framework}} 
& Claude-4.5-Sonnet + UI-Ins-7B  & \underline{43.8} & \underline{26.6} & \underline{0.76} & \underline{0.25} & 1.91 \\
& Gemini-3-Pro + UI-Ins-7B       & 46.3 & \textbf{24.2} & 0.36 & 0.19 & \textbf{2.63} \\
& GPT-5 + UI-Ins-7B              & \textbf{51.7} & 27.8 & \textbf{1.11} & \textbf{0.40} & \underline{2.23} \\
\midrule
\multirow{9}{=}{\centering\textit{End-to-End\\Model}} 
& GUI-Owl-7B                     & 4.5 & \underline{20.6} & - & - & - \\
& GUI-Owl-32B                    & 5.5 & 24.0 & - & - & -     \\
& UI-Venus-7B                    & 5.5 & 26.7 & - & - & - \\
& UI-Venus-72B                   & 10.4 & 34.2 & - & - & - \\
& Qwen3-VL-8B                   & 5.5 & 24.8 & 0.04 & 0.00 & 2.32 \\
& Qwen3-VL-32B                   & 9.0 & 27.1 & 0.00 & 0.00 & 3.84 \\
& Qwen3-VL-235B-A22B             & 9.5 & 26.9 & 0.00 & 0.00 & 2.38 \\
& GELab-Zero-4B       & 10.9 & 29.9 & 0.37 & 0.02 & - \\
& Doubao-1.5-UI-TARS             & 20.9 & 20.9 & 1.22 & 0.13 & - \\
\bottomrule
\end{tabular}
\end{table}

%% file: figures/fig-step-distribution.tex
\begin{wrapfigure}{r}{0.5\textwidth}
\centering
\includegraphics[width=0.45\textwidth]{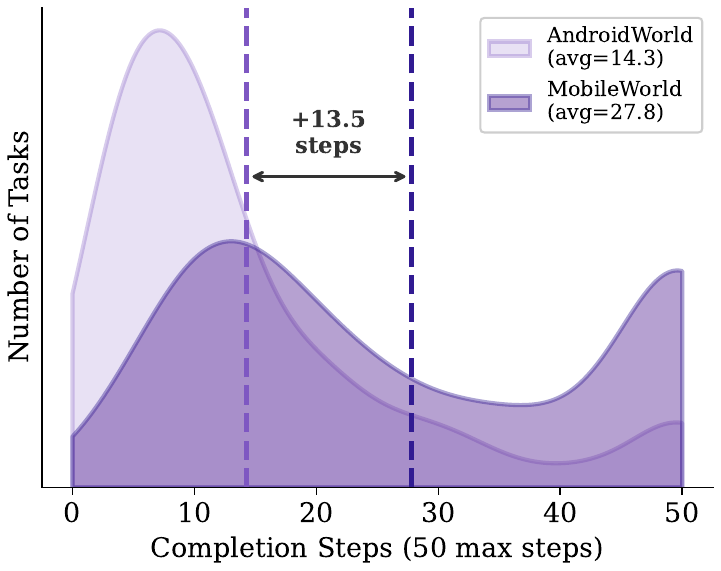}
\vspace{-0.5em}
\caption{Comparison of completion steps between AndroidWorld and \mobileworld{}.}
\label{fig:benchmark_comparison}
\vspace{-2em}
\end{wrapfigure}

%% file: figures/fig-case_study.tex
\begin{figure}[t]
    \centering
    \includegraphics[width=\textwidth,page=1]{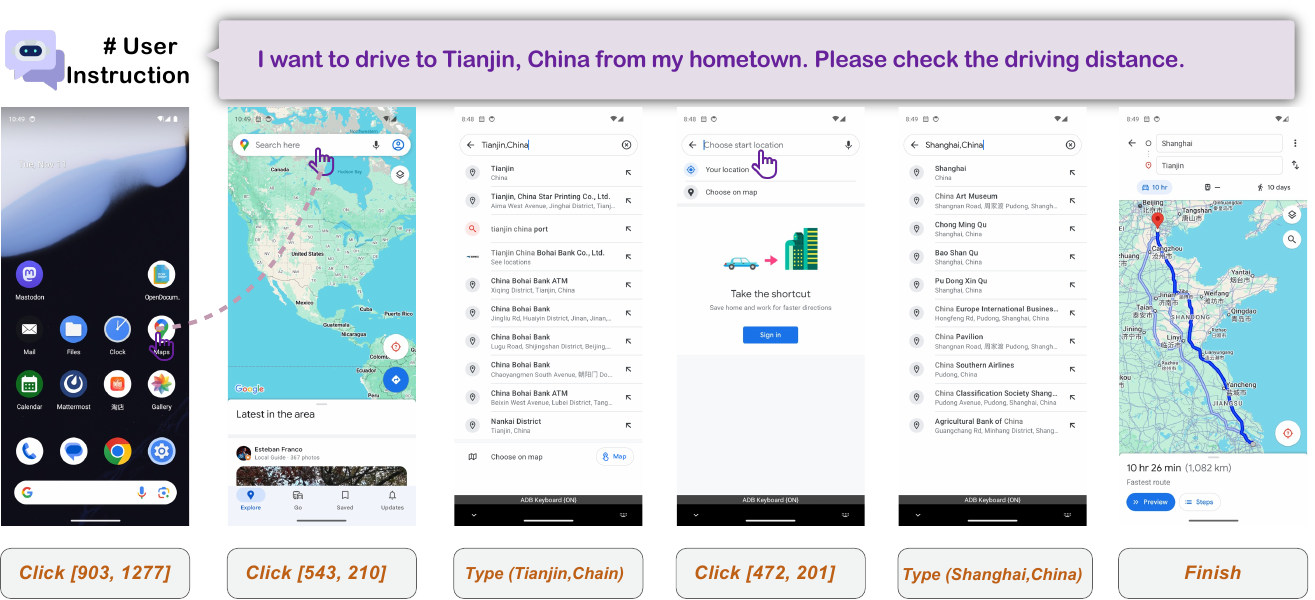}
    \caption{Hallucination without user clarification: A representative failure case showing how the mobile GUI agent hallucinates actions when faced with ambiguous scenarios that require user clarification.}
    \label{fig:hallucination_example}
\end{figure}

\begin{figure}[h]
    \centering
    \includegraphics[width=\textwidth,page=3]{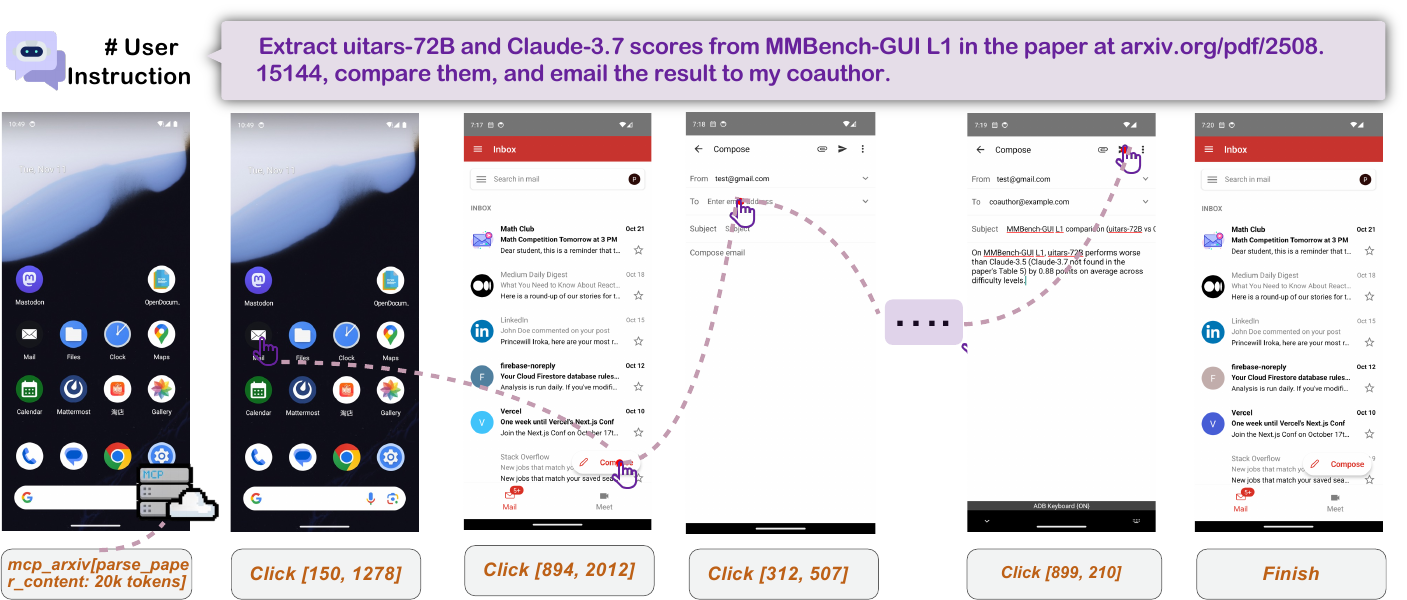}
    \caption{Context overflow from MCP tool responses: A failure case demonstrating ineffective MCP tool integration due to context management issues, where tool responses exceed the context window capacity.}
    \label{fig:mcp_overflow_example}
    
\end{figure}

\begin{figure}[h]
    \centering
    \includegraphics[width=\textwidth,page=2]{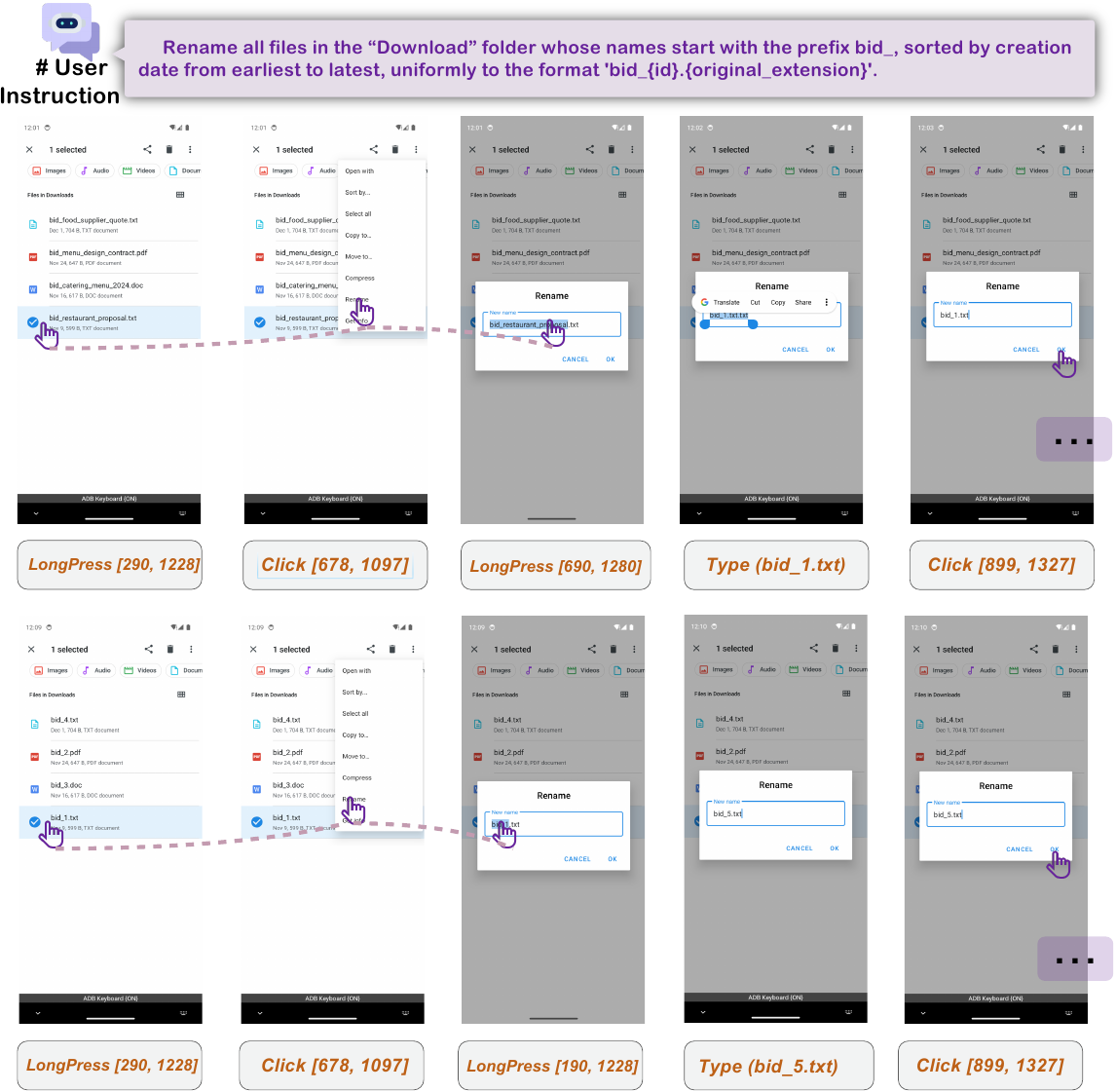}
    \caption{Lack of long-term memory: A representative case illustrating insufficient memory mechanisms for tracking multi-step operations, leading to failure in maintaining state across sequential actions.}
    \label{fig:rename_loop_example}
    
\end{figure}

%% file: content/conclusion.tex
\section{Conclusion}

We introduced \mobileworld{}, a challenging benchmark for evaluating mobile GUI agents that addresses the saturation of existing benchmarks. \mobileworld{} advances mobile agent evaluation through complex workflows, agent-user interaction tasks that assess ambiguity recognition, and MCP-augmented tasks requiring orchestration between GUI operations and external tools. Built upon production-grade open-source applications in a fully containerized environment, our benchmark enables deterministic evaluation through multiple verification methods.

Extensive experiments reveal that state-of-the-art models achieve substantially lower success rates on \mobileworld{} compared to existing benchmarks, with the best agentic framework achieving only 51.7\% success rate and the best end-to-end models reaching just 20.9\%. Performance further degrades on agent-user interaction and MCP-augmented tasks, highlighting critical gaps in ambiguity detection, collaborative dialogue, and hybrid execution planning.

\paragraph{Future Research Directions}
Our failure analysis points to several research directions. We identify two possiblilities here. First, \textbf{foundation model improvements} through reinforcement learning can enhance core reasoning capabilities, numerical computation accuracy, and generalization to long-horizon tasks. Second, \textbf{agentic framework innovations} addressing context length limitation for long-term memories, MCP tool management can bridge the gap between model capabilities and practical deployment requirements.
We hope \mobileworld{} will serve as a rigorous testbed to drive progress toward mobile agents capable of handling real-world automation complexity.

%% file: content/appendix.tex
\newpage

\section{Planner-Executor Agentic Framework Details}
\label{sec:agent_framework_appendix}

\subsection{Planner}
The planner prompt is shown in the code block below. We define an action space that aligns with the one implemented in \mobileworld{}. In addition, we integrate the \texttt{ask\_user} action to enable clarification of ambiguous instructions. Furthermore, when a task is tagged as MCP-augmented, we dynamically inject the specifications of the relevant MCP tools into the prompt.
\input{tables/planner_agent_prompt}

\subsection{Grounding Executor}
For the grounding executor in our agentic framework, we design it to take click-related instructions generated by the upstream planner and output precise screen coordinates for clicking. The prompt used for this module is shown in the code snippet below.
\input{tables/grounding_prompt}

\section{APP Information}

\subsection{APP List}
\Cref{tab:app_info} lists the GUI applications included in \mobileworld{}, along with the number of tasks associated with each. These apps cover a broad range of everyday mobile usage scenarios, including communication (e.g., Mail, Messages), productivity (e.g., Calendar, DocsReader), navigation (Google Map), social interaction (Mastodon, Mattermost), and e-commerce (Taodian). To ensure reproducibility and enable deterministic evaluation, we prioritize open-source or publicly available implementations that closely mimic the functionality of popular commercial applications, such as Taobao (via Taodian), Gmail (via Mail), and Slack (via Mattermost). This design choice allows us to maintain full control over the environment while still providing realistic, real-world-aligned tasks. By using these open-source equivalents, we strike a balance between ecological validity and rigorous, verifiable assessment, enabling fair and repeatable benchmarking of mobile agents.
\input{tables/app_info}

\subsection{Open-source APP Environment Construction}
\label{sec:app_construction}

To ensure reproducibility and enable deterministic evaluation, we construct self-hosted backend environments for key applications. This section details the implementation strategies for our four primary open-source applications.

\paragraph{Mattermost}
We build the Mattermost environment based on the official Docker deployment repository\footnote{\url{https://github.com/mattermost/docker}}, which provides a Docker Compose-based setup for the Mattermost service. Initial chat histories are manually generated following the official import guide using the Mattermost CLI tool. All backend data, including PostgreSQL database contents and file storage, are consolidated into a single directory. To ensure consistent initialization across task executions, we snapshot this directory and restore it at the beginning of each evaluation by copying the contents to the designated location before launching the Docker Compose stack. This approach guarantees that each task starts from an identical initial state. Additionally, we develop auxiliary tools that leverage the CLI to dynamically generate new chat messages during initialization when required by specific tasks. For evaluation, we directly query the PostgreSQL database to verify task outcomes, such as the presence of target messages or the creation of chat groups.

\paragraph{Mastodon}
The Mastodon environment is constructed using the official Docker setup from the Mastodon repository\footnote{\url{https://github.com/mastodon/mastodon}}. We manually create initial posts and user accounts within the platform, then capture a snapshot of the complete backend state, including the PostgreSQL database and media storage. Similar to Mattermost, we employ a snapshot-and-restore strategy: the backend data directory is preserved and restored before each task execution to ensure reproducibility. Since the original Mastodon Android client enforces HTTPS connections, we apply minimal modifications to the client application to enable communication with the locally hosted HTTP backend service. Task verification is performed by directly querying the PostgreSQL database to validate outcomes such as new post creation or user interactions.

\paragraph{Mail App}
We develop the Mail application based on a pure-frontend React Native Gmail clone\footnote{\url{https://github.com/PrincewillIroka/gmail_clone}}. We resolve multiple compilation errors that arose with recent Android SDK versions and extend the application with critical functionalities, including attachment selection, email search capabilities, and email composition via the Android \texttt{Share} interface. To enable programmatic evaluation, we further augment the app with a callback mechanism that persists email sending events to a local file. When an email is sent, relevant metadata (recipient, subject, body, attachments) is recorded and subsequently retrieved by the evaluation function to verify task completion.

\paragraph{Taodian App}
The Taodian e-commerce application is adapted from the Mall4Uni full-stack platform\footnote{\url{https://gitee.com/gz-yami/mall4uni}}. To simplify deployment and eliminate external dependencies, we replace the original backend service with a lightweight file-based mock server. Product catalogs are manually curated from publicly available online sources. We enhance the user interface with improved theming and implement additional features such as SMS-based login and a refined shopping cart management interface. Similar to the Mail app, we instrument the application with callback hooks at critical interaction points (e.g., checkout button clicks) to capture shopping cart contents and shipping details. This information is transmitted to a callback server and subsequently used for automated task validation.

This comprehensive construction approach ensures that all applications are fully self-contained, reproducible, and amenable to automated evaluation while closely approximating real-world mobile application functionality.

 \begin{figure}[t!]
    \centering
    \includegraphics[width=\textwidth]{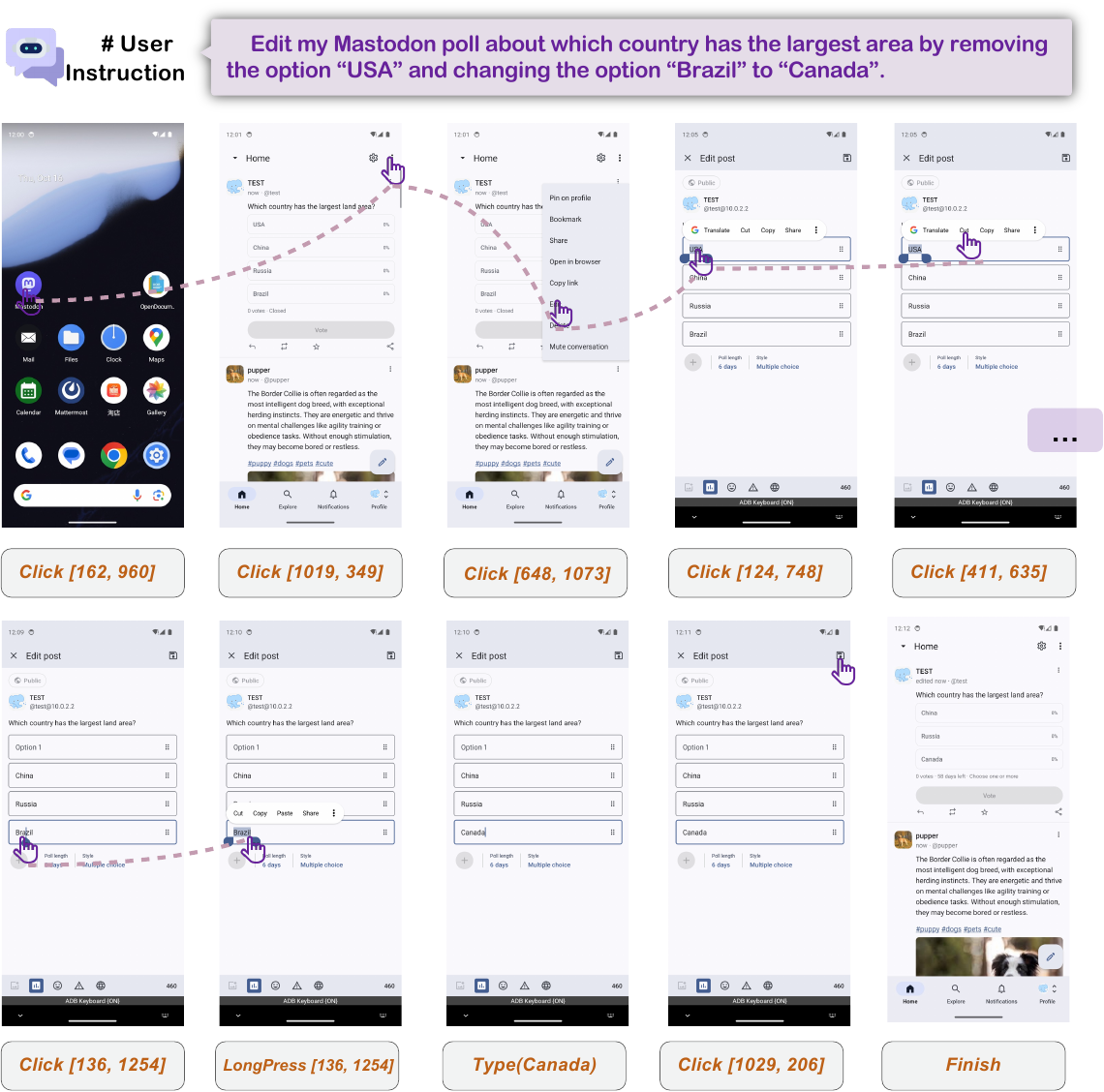}
    \caption{Task Completion Task Example}
    \label{fig:tc_example}
\end{figure}
\section{Case Study}

In this section, we demonstrate various types of examples to illustrate the tasks in \mobileworld{}.

\subsection{GUI-Only Task Example: Task Completion}
\Cref{fig:tc_example} demonstrates editing a Mastodon poll by removing the ``USA'' option and replacing ``Brazil'' with ``Canada'', guided by precise click and input actions. The process involves navigating to the post, accessing the edit menu, deleting the unwanted option, and typing the new one.

\subsection{GUI-Only Task Example: Information Retrieval}

\begin{figure}[h]
    \centering
    \includegraphics[width=\textwidth, page=2]{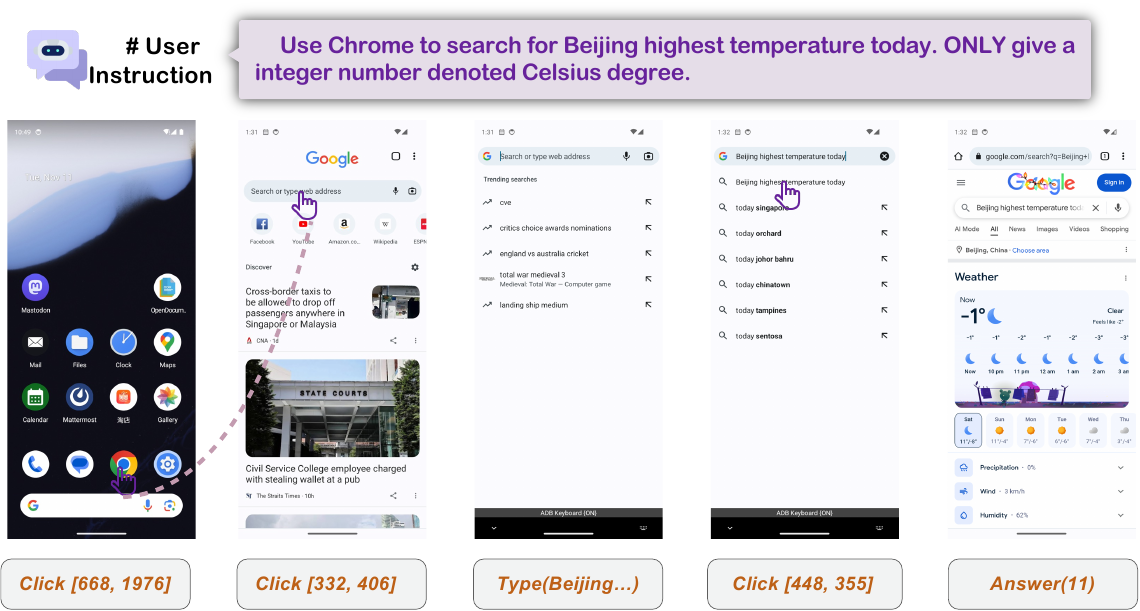}
    \caption{Information Retrieval Task Example}
    \label{fig:ir_example}
\end{figure}

\Cref{fig:ir_example} presents a real-time mobile task in which the user retrieves Beijing's highest temperature for \textit{today} using the Chrome browser. Starting from the home screen, the agent should launch Chrome, inputs the query ``Beijing highest temperature today'', and immediately extracts the current forecast from Google's dynamic weather widget. Both the execution and validation of the task are inherently real-time, ensuring that the final output (11°C) reflects the actual temperature condition on the day of execution.

\subsection{Agent-User Interaction Task Example}
\begin{figure}[t!]
    \centering
    \includegraphics[width=0.75\textwidth, page=3]{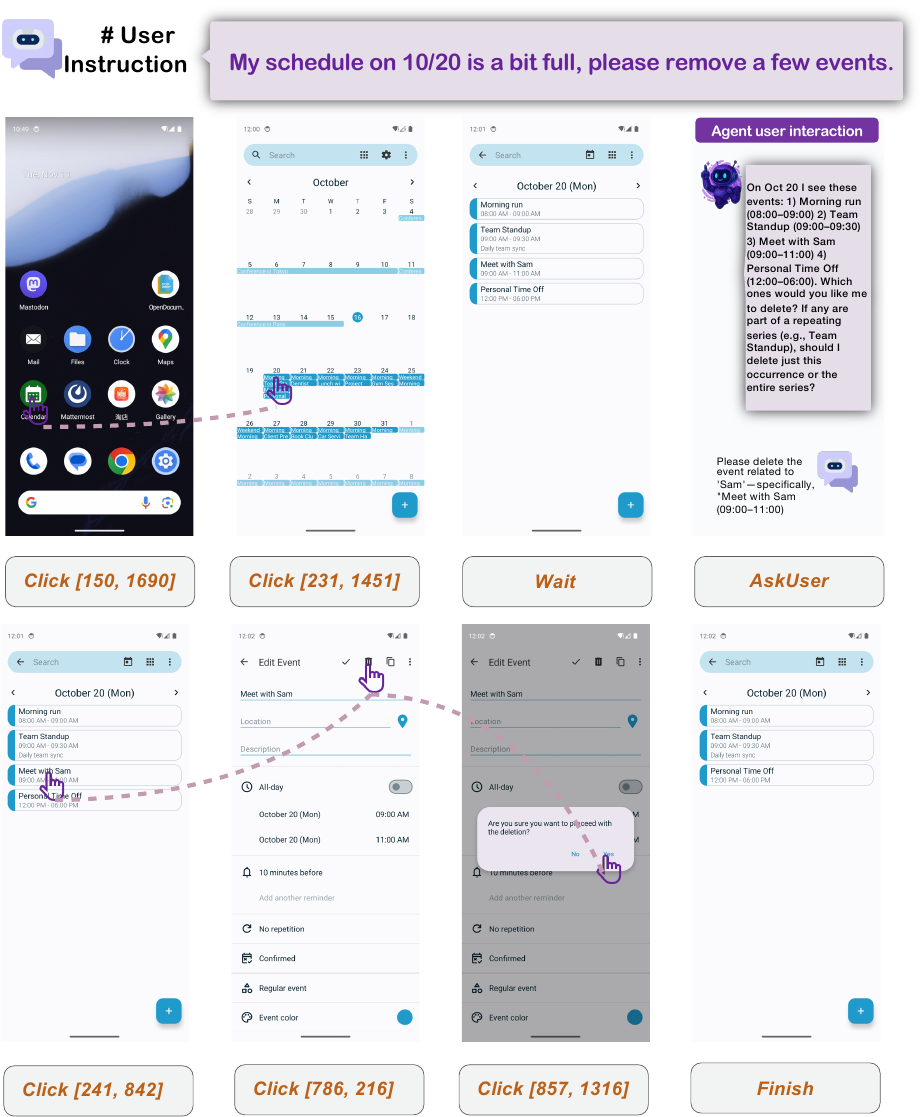}
    \caption{Agent-User Interaction Task Example}
    \label{fig:aui_example}
\end{figure}

\Cref{fig:aui_example} highlights a task involving calendar event management with an ambiguous user instruction: ``My schedule on 10/20 is a bit full, please remove a few events.'' The directive lacks specificity regarding which events to delete, making it inherently ambiguous. To resolve this uncertainty, the agent must engage in interactive clarification. In this scenario, the system identifies multiple events on October 20 and explicitly asks whether to delete only the occurrence of ``Meet with Sam'' or the entire recurring series. This interaction underscores the sensitivity and importance of user confirmation in critical operations such as event deletion, where incorrect actions could disrupt scheduling. 

\newpage
\subsection{MCP-Augmented Task Example}
\begin{figure}[t!]
    \centering
    \includegraphics[width=\textwidth,page=1]{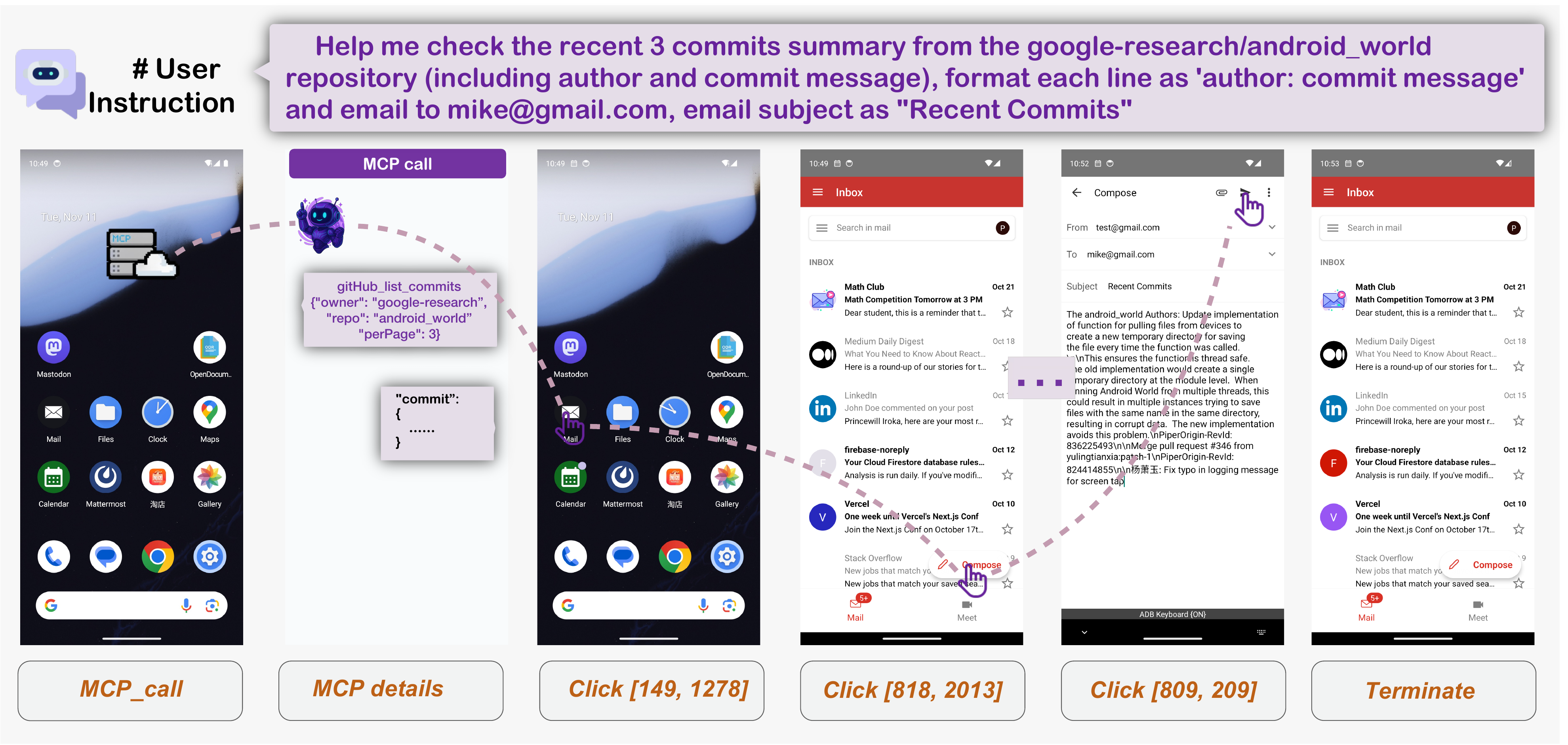}
    \caption{MCP-Augmented Task Example}
    \label{fig:mcp_example}
\end{figure}

The example shown in ~\Cref{fig:mcp_example} demonstrates how a mobile GUI agent completes a complex task with the help of MCP tool invocation: the user requests the three most recent commits from the google-research/android\_world repository, formatted as ``author: commit message,'' and sent via email to a specified address. The agent first invokes an MCP tool to fetch the commit data, then opens the Mail app, taps ``Compose'', and fills in the recipient, subject, and body with the retrieved information before successfully sending the email. The example highlights the agent's ability to seamlessly integrate external API calls with precise UI interactions for end-to-end task execution in real-world mobile environments.

%% file: tables/planner_agent_prompt.tex
\begin{tcolorbox}[title=System Prompt,
breakable,
  fonttitle=\scriptsize\bfseries,
  fontupper=\scriptsize]

\textbf{\#\# Role: Android Phone Operator AI}

You are an AI that controls an Android phone to complete user requests. Your responsibilities:
\begin{itemize}[leftmargin=2.5mm,itemsep=0.2ex,topsep=0.2ex]
  \item Answer questions by retrieving information from the phone.
  \item Perform tasks by executing precise actions.
\end{itemize}

\medskip
\noindent\textbf{\#\# Action Framework.} Respond with \textbf{EXACT JSON} format for one of these actions:\\[-0.25ex]
{\scriptsize
\renewcommand{\arraystretch}{1.2}
\begin{tabularx}{\linewidth}{@{}p{0.18\linewidth} p{0.32\linewidth} >{\raggedright\arraybackslash}X@{}}
\toprule
\textbf{Action} & \textbf{Description} & \textbf{JSON Format Example} \\
\midrule
\texttt{click} & Tap visible element (describe clearly) & {\tiny\ttfamily\raggedright \{ "action\_type": "click", "target": "blue circle button at top-right" \}} \\[0.5ex]
\texttt{double\_tap} & Double-tap visible element (describe clearly) & {\tiny\ttfamily\raggedright \{ "action\_type": "double\_tap", "target": "blue circle button at top-right" \}} \\[0.5ex]
\texttt{long\_press} & Long-press visible element (describe clearly) & {\tiny\ttfamily\raggedright \{ "action\_type": "long\_press", "target": "message from John" \}} \\[0.5ex]
\texttt{drag} & Drag from visible element to another visible element (describe both clearly) & {\tiny\ttfamily\raggedright \{ "action\_type": "drag", "target\_start": "the start point", "target\_end": "the end point" \}} \\[0.5ex]
\texttt{input\_text} & Type into field (includes clicking field, typing, and pressing enter) & {\tiny\ttfamily\raggedright \{ "action\_type":"input\_text", "text":"Hello" \}} \\[0.5ex]
\texttt{answer} & Respond to user & {\tiny\ttfamily\raggedright \{ "action\_type":"answer", "text":"It's 25 degrees today." \}} \\[0.5ex]
\texttt{navigate\_home} & Return to home screen & {\tiny\ttfamily\raggedright \{ "action\_type": "navigate\_home" \}} \\[0.5ex]
\texttt{navigate\_back} & Navigate back & {\tiny\ttfamily\raggedright \{ "action\_type": "navigate\_back" \}} \\[0.5ex]
\texttt{scroll} & Scroll direction (up/down/left/right) & {\tiny\ttfamily\raggedright \{ "action\_type":"scroll", "direction":"down" \}} \\[0.5ex]
\texttt{status} & Mark task as \texttt{complete} or \texttt{infeasible} & {\tiny\ttfamily\raggedright \{ "action\_type":"status", "goal\_status":"complete" \}} \\[0.5ex]
\texttt{wait} & Wait for screen to update & {\tiny\ttfamily\raggedright \{ "action\_type":"wait" \}} \\[0.5ex]
\texttt{ask\_user} & Ask user for information & {\tiny\ttfamily\raggedright \{ "action\_type":"ask\_user", "text":"what is the exact requirements?" \}} \\[0.5ex]
\texttt{keyboard\_enter} & Press enter key & {\tiny\ttfamily\raggedright \{ "action\_type":"keyboard\_enter" \}} \\
\bottomrule
\end{tabularx}
}

\medskip
\textbf{\#\# Execution Principles.}
\begin{enumerate}[leftmargin=2.8mm,itemsep=0.2ex,topsep=0.2ex]
  \item \textbf{Communication Rule:}
    \begin{itemize}[leftmargin=3mm,itemsep=0.1ex,topsep=0.1ex]
      \item ALWAYS use 'answer' action to reply to users - never assume on-screen text is sufficient
      \item Please follow the user instruction strictly to answer the question, e.g., only return a single number, only return True/False, only return items separated by comma.
      \item NEVER use 'answer' action to indicate waiting or loading - use 'wait' action instead
      \item Note that \texttt{answer} will terminate the task immediately.
    \end{itemize}
  \item \textbf{Efficiency First:}
    \begin{itemize}[leftmargin=3mm,itemsep=0.1ex,topsep=0.1ex]
      \item Choose simplest path to complete tasks
      \item If action fails twice, try alternatives (e.g., long\_press instead of click)
    \end{itemize}
  \item \textbf{Smart Navigation:}
    \begin{itemize}[leftmargin=3mm,itemsep=0.1ex,topsep=0.1ex]
      \item Gather information when needed (e.g., open Calendar to check schedule)
      \item For scrolling:
        \begin{itemize}[leftmargin=2mm]
          \item Scroll direction is INVERSE to swipe (scroll down to see lower content)
          \item If scroll fails, try opposite direction
        \end{itemize}
    \end{itemize}
  \item \textbf{Text Operations:}
    \begin{itemize}[leftmargin=3mm,itemsep=0.1ex,topsep=0.1ex]
      \item You MUST first click the input box to activate it before typing the text.
      \item For text manipulation:
        \begin{enumerate}[leftmargin=2mm]
          \item Long-press to select
          \item Use selection bar options (Copy/Paste/Select All)
          \item Delete by selecting then cutting
        \end{enumerate}
    \end{itemize}
  \item \textbf{Ask User:}
    \begin{itemize}[leftmargin=3mm,itemsep=0.1ex,topsep=0.1ex]
      \item If you think you have no enough information to complete the task, you should use \texttt{ask\_user} action to ask the user to get more information.
    \end{itemize}
\end{enumerate}

\medskip
\textbf{\#\# Decision Process.}
\begin{enumerate}[leftmargin=2.8mm,itemsep=0.2ex,topsep=0.2ex]
  \item Analyze goal, history, and current screen
  \item Determine if task is already complete (use \texttt{status} if true)
  \item If not, choose the most appropriate action to complete the task.
  \item Output in exact format below, and ensure the Action is a valid JSON string:
  \item The action output format is different for GUI actions and MCP tool actions. Note only one tool call is allowed in one action.
\end{enumerate}

\medskip
\textbf{\#\# Expected Output Format} (\texttt{Thought:} and \texttt{Action:} are required):

Thought: [Analysis including reference to key steps/points when applicable]\\
Action: [Single JSON action]

\medskip
\textbf{\#\# Output Format Example}

\textit{for GUI actions:}
\begin{tcolorbox}[colback=black!3,colframe=black!10,arc=0.6mm,boxsep=0.6mm,left=0.8mm,right=0.8mm,top=0.6mm,bottom=0.6mm]
{\tiny\ttfamily
Thought: I need to ... to complete the task.\\
Action: \{ "action\_type": "type", "text": "What is weather like in San Francisco today?" \}
}
\end{tcolorbox}

\textit{for MCP tools:}
\begin{tcolorbox}[colback=black!3,colframe=black!10,arc=0.6mm,boxsep=0.6mm,left=0.8mm,right=0.8mm,top=0.6mm,bottom=0.6mm]
{\tiny\ttfamily
Thought: I need to use the provided mcp tool to get the information...\\
Action: \{ "action\_type": "mcp", "action\_json": tool\_args\_obj, "action\_name": "mcp\_tool\_name" \}
}
\end{tcolorbox}

\medskip
\textbf{\#\# Available MCP Tools}

\texttt{\{\{ tools \}\}}

\medskip
\textbf{\#\# User Goal}

\texttt{\{\{ goal \}\}}

\end{tcolorbox}

%% file: tables/grounding_prompt.tex
\begin{tcolorbox}[title=System Prompt,
breakable,
  fonttitle=\scriptsize\bfseries,
  fontupper=\scriptsize]

You are a GUI agent. You are given a task and your action history, with screenshots. You need to perform the next action to complete the task. 
\#\#  Output Format
Return a json object with function name and arguments within \texttt{<tool\_call></tool\_call>} XML tags:

\begin{verbatim}
<tool_call>
{"name": "grounding", "arguments": <args-json-object>}
</tool_call>
\end{verbatim}

\texttt{<args-json-object>} represents the following item of the action space:

\#\# Action Space

\verb|{"action": "click", "coordinate": [x, y]}|

\end{tcolorbox}

%% file: tables/app_info.tex
\begin{table}[h!]
\centering
\small
\caption{List of \mobileworld{} GUI apps and number of tasks for each one.}
\label{tab:app_info}
\begin{tabular}{llcc}
\toprule
\textbf{App} & \textbf{Description} & \textbf{Comparable Commercial App} & \textbf{\#Tasks} \\
\midrule
Calendar & Manage events and schedules & Google Calendar & 30 \\
Camera & Take photos and videos & - & 3 \\
Chrome & Web browser for internet browsing & - & 15 \\
Clock & Alarms, timers, and world clock & - & 7 \\
Contacts & Manage contact information & - & 11 \\
Docreader & View and read documents & Adobe Reader & 10 \\
Files & File manager for storage & - & 32 \\
Gallery & View and manage photos & - & 11 \\
Mail & Email client for messaging & Gmail & 61 \\
Google Map & Navigation and location services & -& 1 \\
Mastodon & Decentralized social network & Twitter/X & 41 \\
Mattermost & Team collaboration and messaging & Slack & 18 \\
Messages & SMS and chat messaging & - & 41 \\
Settings & System configuration & - & 7 \\
Taodian & E-commerce shopping platform & Taobao & 16 \\
\bottomrule
\end{tabular}
\end{table}